\documentclass{article}

\PassOptionsToPackage{numbers, compress}{natbib}

\usepackage[preprint]{neurips_2026}

\usepackage[T1]{fontenc}
\usepackage[utf8]{inputenc}
\usepackage{mathtools}

\usepackage{amssymb,mathrsfs}
\usepackage{amsthm}
\usepackage{bm}
\usepackage{scalerel}
\usepackage{nicefrac}
\usepackage{microtype} 
\usepackage[shortlabels]{enumitem}
\usepackage{graphicx}
\usepackage{epstopdf}
\DeclareGraphicsExtensions{.eps,.png,.jpg,.pdf}

\usepackage{url}
\usepackage{colortbl}
\usepackage{booktabs}
\usepackage{multirow}
\usepackage{colortbl,xcolor}
\usepackage{soul}
\usepackage{xparse,xstring}
\usepackage{calc}
\usepackage{etoolbox}

\makeatletter
\@ifpackageloaded{natbib}{
	\relax
}{
	\usepackage{cite}
}
\makeatother

\usepackage{array}
\newcolumntype{L}[1]{>{\raggedright\let\newline\\\arraybackslash\hspace{0pt}}m{#1}}
\newcolumntype{C}[1]{>{\centering\let\newline\\\arraybackslash\hspace{0pt}}m{#1}}
\newcolumntype{R}[1]{>{\raggedleft\let\newline\\\arraybackslash\hspace{0pt}}m{#1}}

\makeatletter
\let\MYcaption\@makecaption
\makeatother
\usepackage[font=footnotesize]{subcaption}
\makeatletter
\let\@makecaption\MYcaption
\makeatother

\usepackage{glossaries}
\makeatletter
\sfcode`\.1006

\let\oldgls\gls
\let\oldglspl\glspl

\newcommand\fussy@ifnextchar[3]{%
	\let\reserved@d=#1%
	\def\reserved@a{#2}%
	\def\reserved@b{#3}%
	\futurelet\@let@token\fussy@ifnch}
\def\fussy@ifnch{%
	\ifx\@let@token\reserved@d
		\let\reserved@c\reserved@a
	\else
		\let\reserved@c\reserved@b
	\fi
	\reserved@c}

\renewcommand{\gls}[1]{%
\oldgls{#1}\fussy@ifnextchar.{\@checkperiod}{\@}}
\renewcommand{\glspl}[1]{%
\oldglspl{#1}\fussy@ifnextchar.{\@checkperiod}{\@}}

\newcommand{\@checkperiod}[1]{%
	\ifnum\sfcode`\.=\spacefactor\else#1\fi
}

\robustify{\gls}
\robustify{\glspl}
\makeatother

\newacronym{wrt}{w.r.t.}{with respect to}
\newacronym{RHS}{R.H.S.}{right-hand side}
\newacronym{LHS}{L.H.S.}{left-hand side}
\newacronym{iid}{i.i.d.}{independent and identically distributed}
\newacronym{SOTA}{SOTA}{state-of-the-art}

\usepackage{float}

\ifx\notloadhyperref\undefined
	\ifx\loadbibentry\undefined
		\usepackage[hidelinks,hypertexnames=false]{hyperref}
	\else
		\usepackage{bibentry}
		\makeatletter\let\saved@bibitem\@bibitem\makeatother
		\usepackage[hidelinks,hypertexnames=false]{hyperref}
		\makeatletter\let\@bibitem\saved@bibitem\makeatother
	\fi
\else
	\ifx\loadbibentry\undefined
		\relax
	\else
		\usepackage{bibentry}
	\fi
\fi

\usepackage{cleveref-forward}

\crefname{equation}{}{}
\Crefname{equation}{}{}
\crefname{claim}{claim}{claims}
\crefname{step}{step}{steps}
\crefname{line}{line}{lines}
\crefname{condition}{condition}{conditions}
\crefname{dmath}{}{}
\crefname{dseries}{}{}
\crefname{dgroup}{}{}
\crefname{page}{page}{pages}

\crefname{Problem}{Problem}{Problems}
\crefformat{Problem}{Problem~#2#1#3}
\crefrangeformat{Problem}{Problems~#3#1#4 to~#5#2#6}

\crefname{Theorem}{Theorem}{Theorems}
\crefname{Corollary}{Corollary}{Corollaries}
\crefname{Proposition}{Proposition}{Propositions}
\crefname{Lemma}{Lemma}{Lemmas}
\crefname{Definition}{Definition}{Definitions}
\crefname{Example}{Example}{Examples}
\crefname{Assumption}{Assumption}{Assumptions}
\crefname{Remark}{Remark}{Remarks}
\crefname{Rem}{Remark}{Remarks}
\crefname{remarks}{Remarks}{Remarks}
\crefname{Appendix}{Appendix}{Appendices}
\crefname{Supplement}{Supplement}{Supplements}
\crefname{Exercise}{Exercise}{Exercises}
\crefname{TheoremA}{Theorem}{Theorems}
\crefname{CorollaryA}{Corollary}{Corollaries}
\crefname{PropositionA}{Proposition}{Propositions}
\crefname{LemmaA}{Lemma}{Lemmas}
\crefname{DefinitionA}{Definition}{Definitions}
\crefname{ExampleA}{Example}{Examples}
\crefname{RemarkA}{Remark}{Remarks}
\crefname{AssumptionA}{Assumption}{Assumptions}
\crefname{ExerciseA}{Exercise}{Exercises}
\crefname{algorithm}{Algorithm}{Algorithms}
\crefname{figure}{Fig.}{Figs.}
\crefname{table}{Table}{Tables}
\crefname{section}{Section}{Sections}
\crefname{subsection}{Section}{Sections}
\crefname{subsubsection}{Section}{Sections}

\usepackage{algorithm}
\usepackage{algpseudocode}

\ifx\loadbreqn\undefined
	\relax
\else
	\usepackage{breqn}
\fi

\makeatletter
\def\cleartheorem#1{%
    \expandafter\let\csname#1\endcsname\relax
    \expandafter\let\csname c@#1\endcsname\relax
}
\def\clearthms#1{ \@for\tname:=#1\do{\cleartheorem\tname} }
\makeatother

\ifx\renewtheorem\undefined
	\ifx\useTheoremCounter\undefined
		\newtheorem{Theorem}{Theorem}
		\newtheorem{Corollary}{Corollary}
		\newtheorem{Proposition}{Proposition}
		
	\else
		\newtheorem{Theorem}{Theorem}
		
		\newtheorem{Proposition}[Theorem]{Proposition}
	\fi

	\newtheorem{Definition}{Definition}
	
	\newtheorem{Remark}{Remark}

\fi

\theoremstyle{remark}

\theoremstyle{plain}

\newcommand{\qednew}{\nobreak \ifvmode \relax \else
		\ifdim\lastskip<1.5em \hskip-\lastskip
			\hskip1.5em plus0em minus0.5em \fi \nobreak
		\vrule height0.75em width0.5em depth0.25em\fi}



\NewDocumentCommand{\movedownsub}{e{^_}}{%
	\IfNoValueTF{#1}{%
		\IfNoValueF{#2}{^{}}
	}{%
		^{#1}
	}%
	\IfNoValueF{#2}{_{#2}}
}

\let\latexchi\chi
\RenewDocumentCommand{\chi}{}{\latexchi\movedownsub}

\DeclareSymbolFont{bsfletters}{OT1}{cmss}{bx}{n}
\DeclareSymbolFont{ssfletters}{OT1}{cmss}{m}{n}
\DeclareMathSymbol{\bsfGamma}{0}{bsfletters}{'000}
\DeclareMathSymbol{\ssfGamma}{0}{ssfletters}{'000}
\DeclareMathSymbol{\bsfDelta}{0}{bsfletters}{'001}
\DeclareMathSymbol{\ssfDelta}{0}{ssfletters}{'001}
\DeclareMathSymbol{\bsfTheta}{0}{bsfletters}{'002}
\DeclareMathSymbol{\ssfTheta}{0}{ssfletters}{'002}
\DeclareMathSymbol{\bsfLambda}{0}{bsfletters}{'003}
\DeclareMathSymbol{\ssfLambda}{0}{ssfletters}{'003}
\DeclareMathSymbol{\bsfXi}{0}{bsfletters}{'004}
\DeclareMathSymbol{\ssfXi}{0}{ssfletters}{'004}
\DeclareMathSymbol{\bsfPi}{0}{bsfletters}{'005}
\DeclareMathSymbol{\ssfPi}{0}{ssfletters}{'005}
\DeclareMathSymbol{\bsfSigma}{0}{bsfletters}{'006}
\DeclareMathSymbol{\ssfSigma}{0}{ssfletters}{'006}
\DeclareMathSymbol{\bsfUpsilon}{0}{bsfletters}{'007}
\DeclareMathSymbol{\ssfUpsilon}{0}{ssfletters}{'007}
\DeclareMathSymbol{\bsfPhi}{0}{bsfletters}{'010}
\DeclareMathSymbol{\ssfPhi}{0}{ssfletters}{'010}
\DeclareMathSymbol{\bsfPsi}{0}{bsfletters}{'011}
\DeclareMathSymbol{\ssfPsi}{0}{ssfletters}{'011}
\DeclareMathSymbol{\bsfOmega}{0}{bsfletters}{'012}
\DeclareMathSymbol{\ssfOmega}{0}{ssfletters}{'012}

\makeatletter
\newcommand*\rel@kern[1]{\kern#1\dimexpr\macc@kerna}
\newcommand*\widebar[1]{%
  \begingroup
  \def\mathaccent##1##2{%
    \rel@kern{0.8}%
    \overline{\rel@kern{-0.8}\macc@nucleus\rel@kern{0.2}}%
    \rel@kern{-0.2}%
  }%
  \macc@depth\@ne
  \let\math@bgroup\@empty \let\math@egroup\macc@set@skewchar
  \mathsurround\z@ \frozen@everymath{\mathgroup\macc@group\relax}%
  \macc@set@skewchar\relax
  \let\mathaccentV\macc@nested@a
  \macc@nested@a\relax111{#1}%
  \endgroup
}
\makeatother

\DeclareMathOperator{\var}{var}

\DeclareMathOperator{\cov}{cov}

\newcommand{\ifbcdot}[1]{\ifblank{#1}{\cdot}{#1}}

\DeclarePairedDelimiterX\abs[1]{\lvert}{\rvert}{\ifbcdot{#1}}
\DeclarePairedDelimiterX\parens[1]{(}{)}{\ifbcdot{#1}}
\DeclarePairedDelimiterX\brk[1]{[}{]}{\ifbcdot{#1}}
\DeclarePairedDelimiterX\braces[1]{\{}{\}}{\ifbcdot{#1}}
\DeclarePairedDelimiterX\angles[1]{\langle}{\rangle}{\ifblank{#1}{\cdot,\cdot}{#1}}
\DeclarePairedDelimiterX\ip[2]{\langle}{\rangle}{\ifbcdot{#1},\ifbcdot{#2}}
\DeclarePairedDelimiterX\norm[1]{\lVert}{\rVert}{\ifbcdot{#1}}
\DeclarePairedDelimiterX\ceil[1]{\lceil}{\rceil}{\ifbcdot{#1}}
\DeclarePairedDelimiterX\floor[1]{\lfloor}{\rfloor}{\ifbcdot{#1}}

\DeclareFontFamily{U}{matha}{\hyphenchar\font45}
\DeclareFontShape{U}{matha}{m}{n}{
      <5> <6> <7> <8> <9> <10> gen * matha
      <10.95> matha10 <12> <14.4> <17.28> <20.74> <24.88> matha12
      }{}
\DeclareSymbolFont{matha}{U}{matha}{m}{n}
\DeclareFontSubstitution{U}{matha}{m}{n}

\DeclareFontFamily{U}{mathx}{\hyphenchar\font45}
\DeclareFontShape{U}{mathx}{m}{n}{
      <5> <6> <7> <8> <9> <10>
      <10.95> <12> <14.4> <17.28> <20.74> <24.88>
      mathx10
      }{}
\DeclareSymbolFont{mathx}{U}{mathx}{m}{n}
\DeclareFontSubstitution{U}{mathx}{m}{n}

\DeclareMathDelimiter{\vvvert}{0}{matha}{"7E}{mathx}{"17}
\DeclarePairedDelimiterX\vertiii[1]{\vvvert}{\vvvert}{\ifbcdot{#1}}

\DeclarePairedDelimiterXPP\trace[1]{\operatorname{Tr}}{(}{)}{}{\ifbcdot{#1}} 
\DeclarePairedDelimiterXPP\col[1]{\operatorname{col}}{\{}{\}}{}{\ifbcdot{#1}} 
\DeclarePairedDelimiterXPP\row[1]{\operatorname{row}}{\{}{\}}{}{\ifbcdot{#1}} 
\DeclarePairedDelimiterXPP\erf[1]{\operatorname{erf}}{(}{)}{}{\ifbcdot{#1}}
\DeclarePairedDelimiterXPP\erfc[1]{\operatorname{erfc}}{(}{)}{}{\ifbcdot{#1}}
\DeclarePairedDelimiterXPP\KLD[2]{D}{(}{)}{}{\ifbcdot{#1}\, \delimsize\|\, \ifbcdot{#2}} 
\DeclarePairedDelimiterXPP\op[2]{\operatorname{#1}}{(}{)}{}{#2} 

\DeclarePairedDelimiterXPP\indicate[1]{{\bf 1}}{\{}{\}}{}{\ifbcdot{#1}}

\NewDocumentCommand\ofrac{s m}{%
	\IfBooleanTF#1%
	{\dfrac{1}{#2}}%
	{\frac{1}{#2}}%
}
\NewDocumentCommand\ddfrac{s m m}{%
	\IfBooleanTF#1%
	{\dfrac{\mathrm{d} {#2}}{\mathrm{d} {#3}}}%
	{\frac{\mathrm{d} {#2}}{\mathrm{d} {#3}}}%
}
\NewDocumentCommand\ppfrac{s m m}{%
	\IfBooleanTF#1%
	{\dfrac{\partial {#2}}{\partial {#3}}}%
	{\frac{\partial {#2}}{\partial {#3}}}%
}

\newcommand{\setgiven}{:}
\providecommand\given{}

\DeclarePairedDelimiterX\Set[2]\{\}{%
	\if#1:%
		\renewcommand\given{\SetSymbol{:}}%
	\else%
		\renewcommand\given{\SetSymbol[\delimsize]{#1}}%
	\fi%
#2
}

\NewDocumentCommand\set{s O{\setgiven} m}{%
	\IfBooleanTF#1%
	{\Set*{#2}{#3}}%
	{\Set{#2}{#3}}%
}

\NewDocumentCommand{\evalat}{ s O{\big} m e{_^} }{%
\IfBooleanTF{#1}%
{\left. #3 \right|}{#3#2|}%
\IfValueT{#4}{_{#4}}%
\IfValueT{#5}{^{#5}}%
}

\providecommand\given{}
\DeclarePairedDelimiterXPP\cprob[1]{}(){}{
\renewcommand\given{\nonscript\,\delimsize\vert\allowbreak\nonscript\,\mathopen{}}%
#1%
}
\DeclarePairedDelimiterXPP\cexp[1]{}[]{}{
\renewcommand\given{\nonscript\,\delimsize\vert\allowbreak\nonscript\,\mathopen{}}%
#1%
}

\DeclareDocumentCommand \P { s e{_^} d() g } {%
	\mathbb{P}%
	\IfBooleanTF{#1}%
		{
			\IfValueT{#2}{_{#2}}%
			\IfValueT{#3}{^{#3}}%
			\IfValueTF{#5}{\cprob{#4 \given #5}}{\IfValueT{#4}{\cprob{#4}}}%
		}%
		{
			\IfValueT{#2}{_{#2}}%
			\IfValueT{#3}{^{#3}}%
			\IfValueTF{#5}{\cprob*{#4 \given #5}}{\IfValueT{#4}{\cprob*{#4}}}%
		}%
}

\DeclareDocumentCommand \E { s e{_^} o g } {%
	\mathbb{E}%
	\IfBooleanTF{#1}%
		{
			\IfValueT{#2}{_{#2}}%
			\IfValueT{#3}{^{#3}}%
			\IfValueTF{#5}{\cexp{#4 \given #5}}{\IfValueT{#4}{\cexp{#4}}}%
		}%
		{
			\IfValueT{#2}{_{#2}}%
			\IfValueT{#3}{^{#3}}%
			\IfValueTF{#5}{\cexp*{#4 \given #5}}{\IfValueT{#4}{\cexp*{#4}}}%
		}%
}

\DeclareDocumentCommand \Var { s e{_^} d() g } {%
	\var%
	\IfBooleanTF{#1}%
		{
			\IfValueT{#2}{_{#2}}%
			\IfValueT{#3}{^{#3}}%
			\IfValueTF{#5}{\cprob{#4 \given #5}}{\IfValueT{#4}{\cprob{#4}}}%
		}%
		{
			\IfValueT{#2}{_{#2}}%
			\IfValueT{#3}{^{#3}}%
			\IfValueTF{#5}{\cprob*{#4 \given #5}}{\IfValueT{#4}{\cprob*{#4}}}%
		}%
}

\DeclareDocumentCommand \Cov { s e{_^} d() g } {%
	\cov%
	\IfBooleanTF{#1}%
		{
			\IfValueT{#2}{_{#2}}%
			\IfValueT{#3}{^{#3}}%
			\IfValueTF{#5}{\cprob{#4 \given #5}}{\IfValueT{#4}{\cprob{#4}}}%
		}%
		{
			\IfValueT{#2}{_{#2}}%
			\IfValueT{#3}{^{#3}}%
			\IfValueTF{#5}{\cprob*{#4 \given #5}}{\IfValueT{#4}{\cprob*{#4}}}%
		}%
}

\ExplSyntaxOn
\NewDocumentCommand \dist {m o o} {%
\mathrm{#1}\left(%
	\IfValueT{#3}{%
		\tl_if_blank:nTF{ #3 }{\cdot\, \middle|\, }{#3\, \middle|\, }%
	}
	\IfValueT{#2}{#2}%
\right)%
}
\ExplSyntaxOff

\NewDocumentCommand {\cbrace} {t+ D[]{black} D(){\widthof{#5}} m m } {%
	\begingroup%
		\color{#2}
		\IfBooleanTF{#1}{%
			\overbrace{#4}^%
		}{
			\underbrace{#4}_%
		}%
		{\parbox[c]{#3}{\centering\footnotesize{#5}}}%
	\endgroup%
}

\let\oldforall\forall
\renewcommand{\forall}{\oldforall \, }

\let\oldexist\exists
\renewcommand{\exists}{\oldexist \, }

\makeatletter

\newcommand{\rankcolor}[2]{%
	\expandafter\renewcommand\csname #1\endcsname[1]{%
		\ifblank{##1}{%
			{\color{#2} \textbf{#2}}%
		}{%
			\ifmmode
				\textcolor{#2}{\bm{##1}}%
			\else%
				{\color{#2} \textbf{##1}}%
			\fi	
		}%
	}
}

\rankcolor{first}{red}
\rankcolor{second}{blue}
\rankcolor{third}{cyan}
\makeatother

\graphicspath{{./Figures/}{./figures/}}
\DeclareDocumentCommand{\includeCroppedPdf}{ o O{./Figures/} m }{
	\IfFileExists{#2#3-crop.pdf}{}{%
		\immediate\write18{pdfcrop #2#3.pdf #2#3-crop.pdf}}%
	\includegraphics[#1]{#2#3-crop.pdf}
}

\makeatletter
\newcommand*{\addFileDependency}[1]{
  \typeout{(#1)}
  \@addtofilelist{#1}
  \IfFileExists{#1}{}{\typeout{No file #1.}}
}
\makeatother

\definecolor{gray90}{gray}{0.9}
\def\colorlist{red,blue,brown,cyan,darkgray,gray,lightgray,green,lime,magenta,olive,orange,pink,purple,teal,violet,white,yellow}

\makeatletter
\def\startcomment{[}
\ifx\nohighlights\undefined
	\newcommand{\createcolor}[1]{%
			\expandafter\newcommand\csname #1\endcsname[1]{{\color{#1} ##1}}%
	}
	\newcommand{\msout}[1]{\text{\color{green} \st{\ensuremath{#1}}}}
	\newcommand{\del}[1]{{\color{green}\ifmmode \msout{#1}\else\st{#1}\fi}}
\else
	\newcommand{\createcolor}[1]{%
			\expandafter\newcommand\csname #1\endcsname[1]{%
				\noexpandarg%
				\StrChar{##1}{1}[\firstletter]%
				\if\firstletter\startcomment%
					\relax
				\else%
					##1
				\fi
			}%
	}
	\newcommand{\msout}[1]{}
	\newcommand{\del}[1]{}
\fi

\def\@tempa#1,{%
    \ifx\relax#1\relax\else
        \createcolor{#1}%
        \expandafter\@tempa
    \fi
}
\expandafter\@tempa\colorlist,\relax,
\makeatother

\newcommand{\hhide}[1]{}

\ifx\diagnoselabel\undefined
	\relax
\else
	\makeatletter
	\def\@testdef #1#2#3{%
		\def\reserved@a{#3}\expandafter \ifx \csname #1@#2\endcsname
			\reserved@a  \else
			\typeout{^^Jlabel #2 changed:^^J%
				\meaning\reserved@a^^J%
				\expandafter\meaning\csname #1@#2\endcsname^^J}%
			\@tempswatrue \fi}
	\makeatother
\fi

\usepackage[utf8]{inputenc} 
\usepackage[T1]{fontenc}    
\usepackage{hyperref}       
\usepackage{url}            
\usepackage{booktabs}       
\usepackage{amsfonts}       
\usepackage{nicefrac}       
\usepackage{microtype}      
\usepackage{xcolor}         

\usepackage{amsmath}
\usepackage{hyperref} 
\usepackage{cleveref} 
\crefname{equation}{}{} 

\usepackage{tabularx}
\usepackage{adjustbox}
\usepackage{wrapfig}
\usepackage{caption}
\title{LongSpike: Fractional Order Spiking State Space Models for Efficient Long Sequence Learning}

\author{%
  Xinrui He \\
  Wuhan University\\
  \texttt{hexinrui@whu.edu.cn} \\
  \And
  Qiyu Kang\thanks{Corresponding author: \texttt{qiyukang@ustc.edu.cn}.} \\
  University of Science and Technology of China\\
  \texttt{qiyukang@ustc.edu.cn} \\
  \And
  Xuhao Li \\
  Anhui University\\
  \texttt{lixh@ahu.edu.cn} \\
  \And
  Zheng-Jun Zha \\
  University of Science and Technology of China\\
  \texttt{zhazj@ustc.edu.cn} \\
}

\begin{document}

\maketitle

\begin{abstract}
  Spiking Neural Networks (SNNs) are well-regarded for their biological plausibility and energy efficiency in processing sequential data. However, dominant SNN architectures typically rely on \emph{first-order} Ordinary Differential Equations (ODEs) to govern neuronal state transitions. This first-order assumption imposes a ``memoryless'' bottleneck, limiting the model's capacity to capture the complex, long-range dependencies inherent in long-sequence tasks. In this work, we propose \textbf{LongSpike}, a novel SNN framework that integrates fractional-order State-Space Modeling (\emph{f}-SSM) from control theory into the spiking domain. By extending traditional integer-order SSMs to the fractional-calculus regime, LongSpike enables the hierarchical integration of neuronal dynamics with long-memory kernels. To mitigate the computational overhead and parallelization challenges typically associated with fractional operators, we leverage a state-space formulation that supports efficient, parallel training. Empirical evaluations on challenging benchmarks, including Long Range Arena (LRA), large-scale WikiText-103, and Speech Commands, demonstrate that LongSpike outperforms state-of-the-art SNNs in accuracy while preserving sparse synaptic computation. The code is available at \url{https://github.com/xinruihe389-commits/LongSpike}.
\end{abstract}

\section{Introduction}
\label{sec:introduction}
Sequential data are ubiquitous in applications such as natural language processing, speech recognition, and time-series analysis. A central challenge in modeling such data lies in capturing long-range dependencies, where current observations depend on events occurring far in the past. Traditional recurrent neural networks (RNNs) \cite{schuster1997bidirectional} maintain a hidden state that evolves over time, but they often suffer from vanishing gradients and a lack of parallel computation, which limits their ability to model long-term dependencies \cite{pascanu2013difficulty}. Gated architectures such as LSTMs \cite{hochreiter1997long} and GRUs \cite{chung2014empirical} partially alleviate this issue. However, they still rely on strictly sequential computation, making them inefficient for very long sequences.

Spiking Neural Networks (SNNs) represent a class of biologically-grounded architectures \cite{maass1997networks,ghosh2009spiking,lee2016training,wu2018spatio} with significant potential for energy-efficient computing. Despite these advantages, traditional SNNs are also inherently recurrent, as the neural potential state updates strictly follow a sequential order \cite{eshraghian2021training}. Transformers address the limitations of RNNs through self-attention, enabling parallel computation \cite{vaswani2017attention}, but the quadratic complexity of attention restricts scalability to long sequences. Even efficient Transformer variants \cite{wang2020linformer,kitaev2020reformer,beltagy2020longformer} struggle to model fine-grained temporal dynamics over long horizons.

Recently, state-space models (SSMs) have emerged as an effective alternative for long-sequence modeling \cite{gu2020hippo,gu2021efficiently,gu2022parameterization,gu2024mamba,dao2024transformers}. By formulating sequence modeling as a dynamical system, SSMs achieve near linear-time complexity and strong inductive biases for temporal data. The integration of SSMs and SNNs is natural, and recent examples \cite{shen2025spikingssms,du2024spiking,lv2025spikebert} enable long-sequence learning by leveraging the parallel computation of SSMs and the sparse synaptic computation of SNNs simultaneously. However, existing SSMs are fundamentally based on \emph{integer-order} dynamics, which implicitly assume Markovian state transitions. This assumption may still limit their ability to model long-range memory effects and non-local temporal dependencies commonly observed in real-world sequential data \cite{podlubny1998fractional,trigeassou2013infinite}.

In control theory, the fractional-order state-space model (\emph{f}-SSM) framework \cite{matignon1996stability,dzielinski2008stability} provides a principled method for modeling systems with memory, as fractional derivatives depend on the history of a signal through a power-law kernel \cite{kai2004analysis,baleanu2012fractional}. 
Motivated by this property, we propose to model the neural potential state update dynamics using \emph{f}-SSM to capture long-range and non-Markovian temporal dependencies. By extending traditional integer-order SSMs to the fractional-calculus regime, our proposed \textbf{LongSpike} enables the hierarchical integration of neuronal dynamics with long-memory kernels. This is related to recent work \cite{ge2026fractional} that incorporates fractional-order dynamics into neuronal membrane potential charging steps. However, that approach still relies on a strictly recurrent diagram where states are updated iteratively. In this work, to mitigate the computational overhead and parallelization challenges typically associated with fractional operators, we leverage a state-space formulation that supports efficient, parallel training.

\textbf{Main contributions.} Our main contributions are summarized as follows:
\begin{itemize}[label=$\bullet$, topsep=0pt, itemsep=3pt, partopsep=0pt, parsep=0pt,leftmargin=10pt]
\item We introduce LongSpike, a spiking sequence modeling framework that integrates \emph{f}-SSM into SNNs, replacing the standard first-order Markovian state transition with a fractional-order dynamics that captures long-memory effects. We demonstrate how hierarchical integration of long-memory kernels can alleviate the memoryless bottleneck typical of first-order ODE-based SNNs.
\item We address the computational bottleneck of fractional operators by reformulating the dynamics into a parallelizable state-space representation. This formulation allows the model to bypass the constraints of strictly sequential updates, enabling efficient parallel training on long sequences while maintaining the sparse computation inherent in SNNs.
\item We conduct extensive experiments on long-sequence benchmarks, including Long Range Arena (LRA), the large-scale WikiText-103 language modeling dataset, and Speech Commands. Our results show that LongSpike consistently outperforms state-of-the-art SNNs in accuracy and long-range modeling capability while preserving high energy efficiency.
\end{itemize}

\section{Related work}
\subsection{Development of Sequence Models}
Early sequence models such as RNNs \cite{schuster1997bidirectional} and gated variants including LSTMs and GRUs \cite{hochreiter1997long,chung2014empirical} laid the foundation for temporal modeling, but they can suffer from vanishing and exploding gradients, which limits their ability to capture long-range dependencies. Transformers \cite{vaswani2017attention} mitigate this issue through self-attention, enabling parallel computation and improved long-range modeling. However, the quadratic complexity of attention remains a major bottleneck for very long sequences \cite{sun2023retentive,yang2023gated}, motivating more efficient alternatives.

SSMs have recently emerged as a strong paradigm for long-sequence modeling. The HiPPO framework \cite{gu2020hippo} introduced principled state representations based on orthogonal polynomials, while S4 and its variants \cite{gu2021efficiently,gu2022parameterization,gupta2022diagonal} improved efficiency and stability through structured parameterizations. More recent models such as Mamba \cite{gu2024mamba} and Mamba-2 \cite{dao2024transformers} further improve efficiency and selectivity, making SSMs competitive with Transformer-based approaches. Nevertheless, existing SSMs rely on integer-order dynamics and implicitly assume Markovian state transitions, which can limit their ability to represent long-memory effects from the perspective of fractional differential equations (FDEs). 
Specifically, the impulse response of integer-order systems exhibits fast exponential decay $\propto e^{-t / \tau}$, which is characterized as short-tailed. In contrast, the impulse response of an FDE follows a power-law decay $\propto t^{-\alpha}$, providing a long-tailed distribution that inherently preserves fundamental memory over significantly longer horizons.

\subsection{SNNs for Sequential Modeling}
SNNs \cite{maass1997networks,ghosh2009spiking,lee2016training,wu2018spatio} are biologically inspired models that communicate via discrete spikes, enabling event-driven computation and high energy efficiency. Advances in surrogate gradient methods \cite{lee2016training} and Backpropagation Through Time (BPTT) \cite{neftci2019surrogate} have made it feasible to train deep SNNs effectively.  

Recent studies have explored leveraging SNNs for long-sequence modeling by combining spiking dynamics with structured sequence models. Notable examples include spiking variants of SSMs such as SpikingS4 \cite{du2024spiking} and SpikingSSMs \cite{shen2025spikingssms}, as well as hybrid architectures that integrate spiking activations into Transformer- or RWKV-style blocks \cite{lv2025spikebert,zhu2023spikegpt}. Other representative approaches include SpikingLMUFormer \cite{liu2024lmuformer}, based on Legendre Memory Units \cite{voelker2019legendre}, Binary S4D \cite{stan2024learning}, which employs binary spiking activations for improved parallel efficiency, and P-SpikeSSM \cite{bal2025p}, which enhances sparsity through stochastic spiking.  
While these approaches benefit from sparse computation and energy efficiency, their temporal modeling capacity remains limited by integer-order dynamics, leaving long-range dependency modeling as an open challenge.

\subsection{Fractional Calculus in Neural Networks}
Fractional calculus provides a principled framework for modeling systems with memory and non-local temporal dependencies, as fractional derivatives depend on the entire history of a signal through power-law kernels \cite{kai2004analysis,baleanu2012fractional}. FDEs have been successfully applied in physics and engineering to model long-memory processes \cite{podlubny1998fractional,mainardi2022fractional}.
In machine learning, fractional operators have been introduced into neural architectures to enrich temporal and structural modeling \cite{CuiKanLi:C25,KanCuiLi:C25}. Recent studies incorporate fractional calculus into graph learning to improve accuracy and robustness \citep{KanZhaDin:C24,ZhaKanSon:C24}, and into generative modeling through fractional diffusion to encourage more diverse samples \cite{nobis2024generative}.
In spiking neural modeling, fractional dynamics have mainly been explored at the neuron level. For example, fractional leaky integrate-and-fire (LIF) models that improve spike-frequency adaptation and robustness \cite{teka2014neuronal,deng2022fractional}. More recent efforts develop fractional-order SNN frameworks \cite{ge2026fractional}, but these models remain recurrent and require iterative state updates. In contrast, our approach applies fractional dynamics to the state evolution of a state-space model, rather than directly modifying spiking neuron dynamics, which enables scalable long-range sequence modeling with parallel computation.

\section{Preliminaries}
In this section, we will briefly review the key concepts that underpin this work, including fractional-order calculus, FDEs, state-space models, and the LIF neuron model, which form the foundation of our advancements.

\subsection{Basic Fractional Order Calculus}
\label{sec:Fractional Order Calculus}
Fractional-order calculus generalizes the concepts of differentiation and integration to non-integer orders. While integer-order derivatives are local operators, fractional-order derivatives are non-local, making them particularly effective for modeling dynamical systems with hereditary properties and long-range temporal dependencies.

For a differentiable function $y(t)$, the standard first-order derivative is defined by the local rate of change:
$\frac{dy(t)}{dt} = \lim_{\Delta t \to 0} \frac{y(t + \Delta t) - y(t)}{\Delta t}$.
To capture historical dependencies within the derivative itself, we utilize the Caputo formulation, which is widely preferred in physical and engineering applications because it incorporates standard initial conditions.
\begin{Definition}(Caputo Fractional Derivative) \cite{diethelm2010analysis}.
  \label{def:Caputo_fractional_derivative}
  For a function $y(t)$ defined on the interval $[0, T]$, the Caputo fractional derivative of order $\alpha \in (0, 1)$ is defined as:
  \begin{equation}
    \label{eq:Caputo}
    D^\alpha y(t) = \frac{1}{\Gamma(1 - \alpha)} \int_0^t \frac{y'(\tau)}{(t - \tau)^{\alpha}} d\tau,
  \end{equation}
  where $y'(\tau)$ is the first-order derivative of $y(\tau)$, and $\Gamma(\cdot)$ denotes the Gamma function.
\end{Definition}

\begin{Remark}
  The Caputo derivative serves as a bridge between memoryless dynamics and systems with infinite memory. As the order $\alpha$ approaches 1, the integral kernel in \cref{eq:Caputo} simplifies, and the operator converges to the classical first-order derivative. For $0 < \alpha < 1$, the derivative at time $t$ becomes a weighted accumulation of the entire history of $y'(\tau)$, where the power-law kernel $(t - \tau)^{-\alpha}$ governs the rate of decay for past information.
\end{Remark}

\subsubsection{Fractional Differential Equations}
A fractional differential equation replaces the integer-order derivative in a standard ODE with a fractional derivative, which introduces history dependence and can better represent long-memory dynamics.
For comparison, a first-order initial value problem is
\begin{equation}
\frac{dy(t)}{dt} = f(t, y(t)), \quad y(0) = y_0.
\end{equation}
Using the Caputo derivative in Equation~\cref{eq:Caputo}, the corresponding fractional-order form is
\begin{equation}
\label{eq:FDE}
D^{\alpha} y(t) = f(t, y(t)), \quad y(0) = y_0.
\end{equation}
Because closed-form solutions for FDEs are rarely tractable, these systems are typically addressed through numerical methods. Common techniques involve temporal discretization to approximate the non-local fractional operator, such as the fractional Adams–Bashforth–Moulton (ABM) predictor-corrector method \cite{kai2004analysis,KanZhaDin:C24}.

\subsection{State Space Models}
SSMs provide a robust framework for representing linear dynamical systems. These models describe the evolution of a system's internal state over time as a function of its prior state and an external input. In a continuous-time setting, the dynamics are governed by a first-order differential equation.

\begin{Definition}(Continuous-Time State Space Models) \cite{gu2021efficiently,gu2024mamba}.
Let $h(t) \in \mathbb{R}^{N}$ denote the latent state vector and $x(t) \in \mathbb{R}$ denote the input signal at time $t$. The state evolution and the corresponding output $y(t)$ are defined as:
\begin{equation}
    \label{eq:state}
    \frac{dh(t)}{dt} = A h(t) + B x(t),\quad y(t) = C h(t) + D x(t)
\end{equation}

where $A \in \mathbb{R}^{N \times N}$ is the state transition matrix, $B \in \mathbb{R}^{N \times 1}$ is the input matrix, $C \in \mathbb{R}^{1 \times N}$ is the output matrix, and $D \in \mathbb{R}$ is the feedthrough term. 
\end{Definition}

In alignment with existing literature on deep SSMs \cite{gu2022parameterization,gu2024mamba}, we assume $D = 0$ to simplify notations, as the feedthrough term functions as a standard skip-connection. 

To process discrete sequences, the continuous system must be discretized using a sampling step $\Delta$. The resulting Discrete-Time SSM is defined as:
\begin{equation}
    h[t] = \bar{A}h[t-1] + \bar{B}x[t],\quad y[t] = C h[t]
\end{equation}
where $\bar{A}$ and $\bar{B}$ are discretized matrices derived from $A$, $B$, and $\Delta$ via rules such as the Bilinear or Zero-Order Hold (ZOH) transformation \cite{gu2021efficiently}. Structured parameterizations, such as enforcing $\bar{A}$ to be diagonal or low-rank, are often employed to ensure computational efficiency \cite{gupta2022diagonal}.

For efficient training on long sequences, the linear and time-invariant nature of the discretized SSM allows it to be reformulated as a global convolution:
\begin{equation}
    \label{eq:conv1}
    y[k] = \sum_{j=1}^k C\bar{A}^{k-j}\bar{B}x[j]
\end{equation}
By pre-computing the convolution kernel $K = \left( C\bar{B}, C\bar{A}\bar{B}, \dots, C\bar{A}^{L-1}\bar{B} \right)$, where $L$ is the sequence length, the output can be computed as:
$y = x \ast K$.
This representation enables full parallelization across the temporal dimension. By utilizing the Fast Fourier Transform (FFT), the computational complexity is reduced to $O(L \log L)$ \cite{fu2022hungry}. Furthermore, on modern parallel hardware such as GPUs, the kernel $K$ can be computed efficiently through optimized parallel scan or summation algorithms, bypassing the bottlenecks associated with strictly sequential updates.
\begin{figure*}[ht]
  \vskip -0.2in
  \begin{center}
    \centerline{\includegraphics[width=\columnwidth]{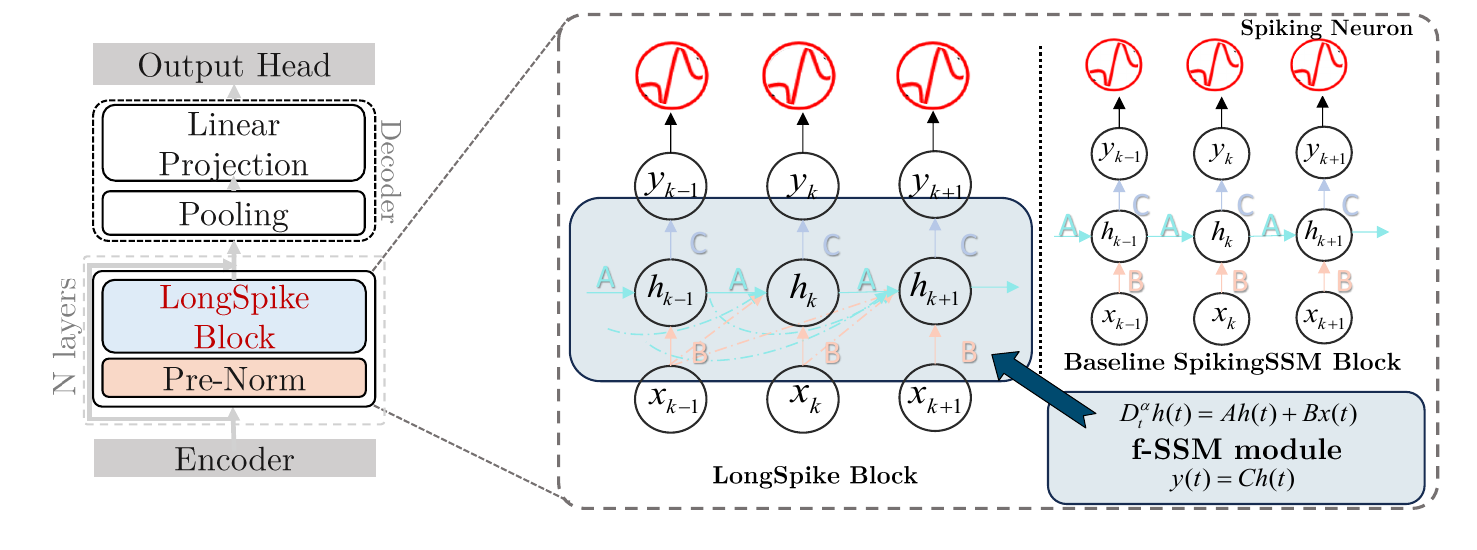}}
    \caption{Overview of the LongSpike architecture. 
    Compared with the baseline SpikingSSM \cite{shen2025spikingssms}, LongSpike incorporates fractional-order state-space dynamics into spiking neurons via an \emph{f}-SSM module, enabling enhanced temporal modeling with efficient parallel computation.
    }
    \label{model}
  \end{center}
  \vskip -0.2in
\end{figure*}
\subsection{LIF Neuron and SNN}
The LIF model is a standard representation of neuronal dynamics in SNNs \cite{stein1967some}. It describes a neuron integrating input signals into a membrane potential $u[t]$, which triggers a spike upon reaching a threshold $v_{\text{th}}$. In discrete time, the potential evolution is governed by:
\begin{equation}
     \bar{u}[t] = \beta u[t-1] + I[t],
\end{equation}
where $\bar{u}[t]$ is the potential before spike determination, $\beta \in (0, 1)$ is the decay constant, and $I[t]$ is the input current. 
The firing state $s[t]$ is determined by the activation function $H(\cdot)$, expressed as:
\begin{equation}
    s[t] = H(\bar{u}[t] - v_{\text{th}}),
\end{equation}
where $v_{\text{th}}$ is the threshold potential and $H$ denotes the Heaviside step function. When the membrane potential reaches this threshold, the neuron generates a spike and the potential is reset. Common reset mechanisms include the soft reset, where the membrane potential is reduced by the threshold value as $u[t] = \bar{u}[t] - s[t] v_{\text{th}}$, and the hard reset, where the potential is set to a fixed value $u_r$ such that $u[t] = \bar{u}[t](1 - s[t]) + s[t] u_r$. In LIF-based SNNs, the sparsity of the resulting spike train $s[t]$ allows for event-driven computation, which significantly reduces energy consumption compared to dense artificial neural networks.

\section{LongSpike Framework}
In this section, we present the LongSpike framework in detail. We begin by formulating the \emph{f}-SSM block, then derive its discretization and parallelizable representation, and finally describe how it integrates with spiking neural dynamics to form the complete LongSpike architecture. By incorporating \emph{f}-SSM, LongSpike replaces standard first-order Markovian state transitions with fractional-order dynamics, enabling the explicit inclusion of long-memory kernels within the spiking domain.

\subsection{Fractional State Space Model Block}
Let $\alpha \in (0, 1)$ denote the fractional order. The \emph{f}-SSM is defined as \cite{matignon1996stability}:
\begin{equation}
    \label{eq:FDE states}
    D^\alpha h(t) = A h(t) + B x(t), \quad y(t) = C h(t),
\end{equation}
where $D^\alpha$ denotes the Caputo fractional derivative, and we assume the initial condition $h(0)=0$. For the state matrix $A$, we adopt the diagonal HiPPO-D parameterization \cite{gu2022parameterization}, which provides a structured initialization for modeling long-range dependencies.

To solve \cref{eq:FDE states}, we transform the differential form into a Volterra integral equation \cite{diethelm2010analysis} by introducing the fractional kernel $k_\alpha(t) = \frac{t^{\alpha - 1}}{\Gamma(\alpha)}$:
\begin{equation}
    \label{eq:integral}
    h(t) = \int_0^t k_\alpha(t - \tau) (A h(\tau) + B x(\tau)) d\tau.
\end{equation}
This integral representation explicitly demonstrates that the state $h(t)$ depends on the complete history of the system's previous states and inputs, weighted by the power-law kernel $k_\alpha$.
While standard numerical solvers (cf. Appendix~\ref{sec:numerical_methods}) can be used for these equations \cite{diethelm2010analysis,zhao2024distributed}, they often lack the parallelization properties required for efficient SSM training. 

To leverage the computational advantages of state-space modeling, we adopt a parallel solving method based on a diffusive representation \cite{jiang2017fast}.
We express the power-law kernel $k_\alpha(t)$ as a mixture of exponentials. Using the identity $\int_0^\infty e^{-\omega t} \omega^{-\alpha} d\omega = \Gamma(1-\alpha)t^{\alpha - 1}$ and the reflection formula $\Gamma(\alpha)\Gamma(1-\alpha) = \pi/\sin(\pi \alpha)$, the kernel can be rewritten as:
\begin{equation}
    \label{eq:diffusive representation}
    k_\alpha(t) = \frac{t^{\alpha - 1}}{\Gamma(\alpha)} = \frac{\sin(\pi \alpha)}{\pi} \int_0^\infty e^{-\omega t} \omega^{-\alpha} d\omega.
\end{equation}
This formulation reveals $k_\alpha(t)$ as an infinite continuum of exponentials. To make this computationally tractable, we approximate the continuum with a finite Sum-of-Exponentials (SOE). By selecting nodes $\{\omega_i\}_{i=1}^{M} \in (0, \infty)$ and weights $\{\eta_i\}_{i=1}^{M}$ that approximate the kernel over the range $t \in [0,T]$, we obtain:
\begin{equation}
    \label{eq:SOE}
    k_\alpha(t) \approx \sum_{i=1}^{M} \eta_i e^{-\omega_i t}.
\end{equation}
Practical methods for generating these $\{\omega_i, \eta_i\}$ pairs are detailed in \cref{sec:Methods for Generating omega&eta} of the Appendix. Furthermore, the $\{\omega_i, \eta_i\}$ pairs may be set as learnable in our implementation.

Subsequently, we define the auxiliary memory states $z_i(t) \in \mathbb{R}^N$ for $i = 1, \dots, M$ as:
\begin{equation}
    z_i(t) = \int_0^t e^{-\omega_i(t - \tau)} g(\tau) d\tau,
\end{equation}
where $g(t) = A h(t) + B x(t)$. This integral is equivalent to the system of first-order linear ODEs:
\begin{equation}
    \dot{z}_i(t) = -\omega_i z_i(t) + g(t), \quad z_i(0) = 0.
\end{equation}
The original state $h(t)$ and output $y(t)$ are then reconstructed via the weighted sum:
\begin{equation}
    \label{eq:reconstructed}
    h(t) \approx \sum_{i=1}^{M} \eta_i z_i(t), \quad y(t) = C h(t).
\end{equation}
Intuitively, this decomposition maps the singular fractional kernel onto a \emph{multi-timescale} memory bank, where $\omega_i$ governs the decay rate and $\eta_i$ acts as a mixing coefficient. Upon discretization (e.g., via ZOH or bilinear transform \cite{gu2022parameterization}), the dynamics of each trace follow:
\begin{equation}
    \label{eq:zi_dt}
    z_i[t] = \bar{a}_i z_i[t-1] + \bar{b}_i g[t],
\end{equation}
where $\bar{a}_i$ and $\bar{b}_i$ are the discretized parameters. Since each trace \eqref{eq:zi_dt} corresponds to a Linear Time-Invariant (LTI) system with an exponential kernel $k_i[\ell]=\bar{b}_i \bar{a}_i^{\ell}$, the full state evolution $z_i = g * k_i$ can be computed via FFT-based convolution. This yields a computational complexity of $O(L\log L)$, aligning with modern efficient SSMs \cite{gu2021efficiently,gu2024mamba}. Thus, LongSpike captures power-law memory without explicit history accumulation, ensuring the computational cost scales linearly with $M$. 

Overall, we avoid explicit power-law history accumulation while preserving long-memory behavior through the SOE approximation.
The cost scales linearly in $M$ (typically small) and inherits the parallel training advantages of SSM-based sequence models.
To further optimize memory efficiency on parallel GPU hardware, we can compute the \emph{effective kernel} $K_{\text{eff}} = \sum_{i=1}^M \eta_i k_i$ via aggregation prior to convolution.
This approach allows us to compute the final state contribution without ever materializing the $M$ intermediate traces $z_i$, significantly reducing memory I/O.

\subsection{LongSpike Block}
The LongSpike architecture integrates fractional state-space dynamics with LIF neurons. The output $y[t]$ from the preceding $f$-SSM block acts as the injection current driving the neuron's membrane potential:
\begin{equation}
    \label{eq:lif_dynamics}
    \bar{u}[t] = \beta u[t-1] + y[t], \quad s[t] = H(\bar{u}[t] - v_{th}),
\end{equation}
where $H(\cdot)$ is the Heaviside step function, $\beta \in (0,1)$ is the membrane decay constant, and $v_{th}$ is a learnable threshold. This coupling enables the model to encode temporal information into sparse, energy-efficient spike representations.

To circumvent the sequential bottleneck inherent in iterative LIF simulation, we adopt a parallelization strategy utilizing a Surrogate Dynamic Network (SDN) \cite{shen2025spikingssms}. The SDN serves as a non-autoregressive mapping that predicts the entire spike train $s_{1:L}$ directly from the input current sequence $y_{1:L}$:
\begin{equation}
    s_{1:L} = f_{\text{SDN}}(y_{1:L}).
\end{equation}
We instantiate the SDN as a CNN layer. This design is theoretically grounded in the observation that the linear recurrence of the LIF dynamics can be effectively approximated by convolutional operators with local receptive fields. During inference, the SDN replaces step-by-step integration with parallel prediction, reducing the sequential complexity to $O(1)$ (constant parallel depth) and significantly enhancing computational throughput.

The full LongSpike architecture is shown in \cref{model}, where the model 
stacks multiple LongSpike blocks on top of an encoder, followed by a 
decoder and an output head.

\subsection{Theoretical Analysis: Long-Memory Properties}
In this section, we theoretically analyze the memory characteristics of the \emph{f}-SSM module relative to standard SSMs. We demonstrate that while both systems converge to identical steady states under constant excitation, \emph{f}-SSM exhibits Mittag–Leffler relaxation characterized by power-law decay. This reveals a persistent, heavy-tailed temporal dependence that is structurally absent in standard first-order models.

\begin{minipage}{0.52\textwidth}
\vspace{-5pt}
\begin{Proposition}[Long-Memory Dynamics]
\label{prop:long_memory}
Consider a scalar state-space model governed by parameters $A = -\lambda$ (with $\lambda > 0$) and $B = 1$, subject to a constant input $x(t) = x_c$ and initial condition $h(0) = 0$. Let the system evolve according to a derivative of order $\alpha \in (0, 1]$. The state trajectories $h(t)$ for the standard ($\alpha=1$) and fractional ($0 < \alpha < 1$) cases are given by:
\begin{align}
    \label{eq:Standard_SSM}
    {\textup{Standard SSM:}} \quad & h(t) = \frac{x_c}{\lambda}  - \frac{x_c}{\lambda} e^{-\lambda t}, \\
    \label{eq:Fractional_SSM}
    {\textup{\emph{f}-SSM:}} \quad & h(t) = \frac{x_c}{\lambda} - \frac{x_c}{\lambda} E_\alpha\left(-\lambda t^\alpha\right),
\end{align}
where $E_\alpha(z) = \sum_{k=0}^\infty \frac{z^k}{\Gamma(\alpha k + 1)}$ is the Mittag–Leffler function \cite{diethelm2010analysis}.
\end{Proposition}
\end{minipage}
\hfill
\begin{minipage}[c]{0.46\textwidth}
\centering
\includegraphics[width=\linewidth]{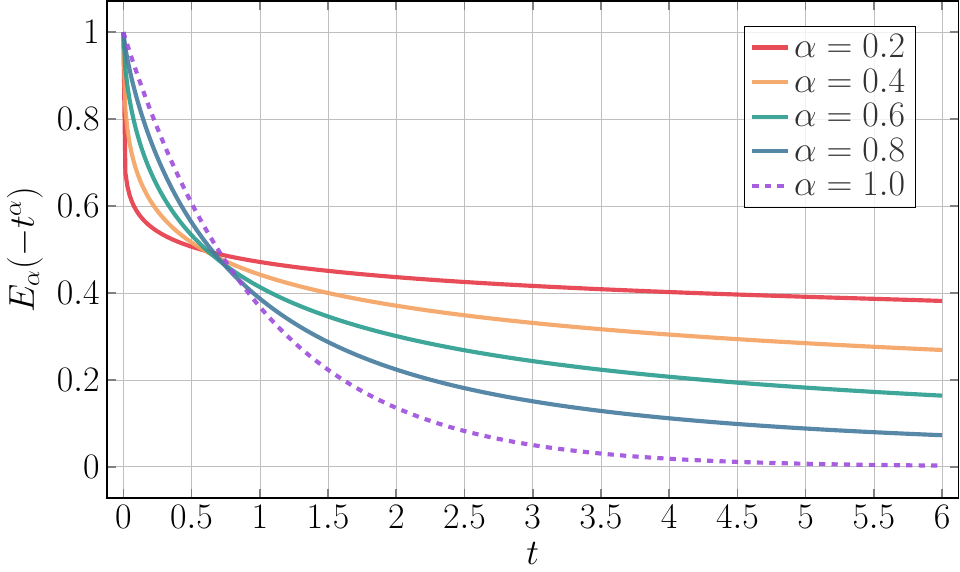}
\captionof{figure}{The Mittag-Leffler function $E_\alpha(-\lambda t^\alpha)$ for varying $\alpha$. For $\alpha=1$ (purple dashed), the function recovers the exponential decay $e^{-\lambda t}$. For $\alpha<1$, the function transitions to an algebraic decay tail, where lower $\alpha$ values indicate stronger long-memory retention.}
\label{fig:mittag_leffler}
\end{minipage}

The asymptotic behavior of the Mittag–Leffler function, illustrated in \cref{fig:mittag_leffler}, highlights the fundamental difference in memory retention:
\begin{itemize}[label=$\bullet$, topsep=0pt, itemsep=3pt, partopsep=0pt, parsep=0pt,leftmargin=20pt]
    \item {Exponential Decay ($\alpha=1$):} The function simplifies to $E_1(z) = e^z$, recovering the fast, memoryless decay of classical systems.
    \item {Algebraic Decay ($\alpha < 1$):} For large $t$, the function exhibits a power-law tail:
    \begin{equation}
        E_\alpha(-\lambda t^\alpha) \sim \frac{1}{\lambda \Gamma(1-\alpha) t^\alpha}, \quad \text{as } t \to \infty.
    \end{equation}
\end{itemize}
Thus, the memory trace in \emph{f}-SSM decays algebraically, significantly slower than the exponential rate of standard SSMs.

\begin{Remark}[Mechanism of Persistence]
While both systems share the same equilibrium point $h_\infty = x_c / \lambda$, their transient relaxation reflects distinct underlying mechanisms. Standard SSMs are Markovian and the influence of past states vanishes exponentially. In contrast, the \emph{f}-SSM is intrinsically non-Markovian due to the non-local nature of the Caputo fractional derivative. The associated history-dependent kernel induces Mittag–Leffler relaxation with heavy tails, ensuring that information from the distant past exerts a non-negligible influence on the current state, a critical property for capturing long-range dependencies in sequence modeling.
\end{Remark}

\section{Experiments}
We evaluate LongSpike on a diverse range of sequence modeling tasks, including the Long Range Arena (LRA) benchmark, large-scale language modeling on WikiText-103, 
and sequential audio classification on Speech Commands. 
Additional experimental results are provided in Appendix~\ref{sec:appendix_extra}. Details of experimental settings are provided in Appendix~\ref{sec:Experimental Settings}.

In addition to benchmark results, we analyze the role of fractional dynamics and the efficiency of LongSpike in terms of computational complexity and energy-related behavior (see \cref{sec:efficiency_analysis}).

\subsection{Datasets}

\textbf{LRA.}  
The Long-Range Arena (LRA) benchmark~\cite{tay2020long} evaluates models on long-context sequences, ranging from 1K to 16K steps, across diverse modalities including text, visual data, and mathematical expressions. It comprises six tasks such as text classification, document retrieval, image classification, pathfinding, and list operations, assessing a model's capacity to capture long-range dependencies and process extended sequences efficiently.

\textbf{WikiText-103.}  
WikiText-103~\cite{merity2016pointer} is a large-scale language modeling dataset containing over 100 million tokens from high-quality Wikipedia articles. Its long, coherent documents make it a standard benchmark for evaluating long-range dependency modeling. We use perplexity (PPL) as the evaluation metric, and this task serves to assess the effectiveness of LongSpike on real-world long-sequence modeling.

\textbf{Speech Commands.}  
The Speech Commands dataset~\cite{warden2018speech} contains over 100,000 one-second audio clips of 35 spoken English words (e.g., "yes," "no," "up," "down") from thousands of speakers, sampled at 16kHz. The task is to classify the spoken words, requiring the model to capture temporal dependencies and variations in speech patterns such as speed, tone, and background noise.

\subsection{Experimental Results}
\label{sec:Experimental Results}

\begin{table*}[!htp]\small
\vspace{-.4cm}
\caption{Performance comparison of LongSpike with other works on LRA benchmark. We take the 50\% accuracy for the absence of Path-X accuracy as did in the work of S4D, then compute the overall average metrics across all tasks as AVG.}
\label{tab:lra_comparison}
\centering
 \resizebox{\textwidth}{!}{
 \setlength{\tabcolsep}{2pt} 
 \begin{tabular}{lcccccccccc}
\toprule
Model & SNN & LISTOPS & TEXT & RETRIEVAL & IMAGE & PATHFINDER & Path-X & AVG \\
\midrule
Transformer & No & 36.37 & 64.27 & 57.46 & 42.44 & 71.40 & - & 53.66 \\
LMUFormer & No & 34.43 & 68.27 & 78.65 & 54.16 & 69.90 & - & 59.24 \\
S4 & No & 58.35 & 76.02 & 87.09 & 87.26 & 86.05 & 88.10 & 80.48 \\
S4D-Lin & No & \textbf{60.52} & \textbf{86.97} & \textbf{90.96} & \textbf{87.93} & \textbf{93.96} & \textbf{92.8} & \textbf{85.52} \\
\midrule
Spiking LMUFormer & Yes & 37.30 & 65.80 & 79.76 & 55.65 & 72.68 & - & 60.20 \\
Binary S4D & Yes & 54.80 & 82.50 & 85.03 & 82.00 & 82.60 & 61.20 & 74.69 \\
P-SpikeSSM & Yes & 58.20 & 81.20 & 88.53 & 82.40 & 84.80 & - & 74.18 \\
SpikingSSM & Yes & 60.23 & 80.41 & 88.77 & 88.21 & 93.51 & 94.82 & 84.33 \\
\textbf{LongSpike} & Yes & \textbf{60.95} & \textbf{88.19} & \textbf{89.82} & \textbf{89.28} & \textbf{94.31} & \textbf{95.41} & \textbf{86.33} \\
\bottomrule
\end{tabular}}
\vspace{-.2cm}
\end{table*}

We first evaluate LongSpike on the six tasks of the LRA benchmark. As shown in Table~\ref{tab:lra_comparison}, it achieves an average score of 85.28\%, outperforming all baseline models. The model is particularly strong on TEXT and RETRIEVAL tasks, demonstrating its ability to capture long-range dependencies efficiently. For illustration, Figure~\ref{metrics_plot} shows the loss and accuracy curves for the RETRIEVAL task, where LongSpike consistently surpasses competing models. It also remains highly competitive on more complex tasks such as PATH-X, highlighting its robustness across diverse sequence modeling challenges. To better understand the source of these improvements, we further evaluate a vanilla (non-spiking) fractional SSM. As shown in Table~\ref{tab:frac_ssm} in the appendix, it consistently outperforms its integer-order counterpart.

\vspace{-5pt}
\begin{minipage}[c]{0.45\textwidth}
\vspace{-5pt}
\centering
\setlength{\tabcolsep}{2pt}
\captionof{table}{Performance comparison on WikiText-103.}
\label{tab:wikitext_comparison}
\begin{tabular}{lcccc}
\toprule
Model & SNN & PPL & Params &Spiking Rate\\
\midrule
Transformer & No & 20.51 & 231M &- \\
S4 & No & 20.95 & 249M &- \\
\midrule
SpikeGPT & Yes & 39.75 & 213M & - \\
SpikingSSM & Yes & 33.94 & 75M & 26.40\% \\
\textbf{LongSpike} & Yes & \textbf{32.31} & 75M &27.05\% \\
\bottomrule
\end{tabular}
\end{minipage}
\hfill
\begin{minipage}[c]{0.48\textwidth}
\centering
\includegraphics[width=\linewidth]{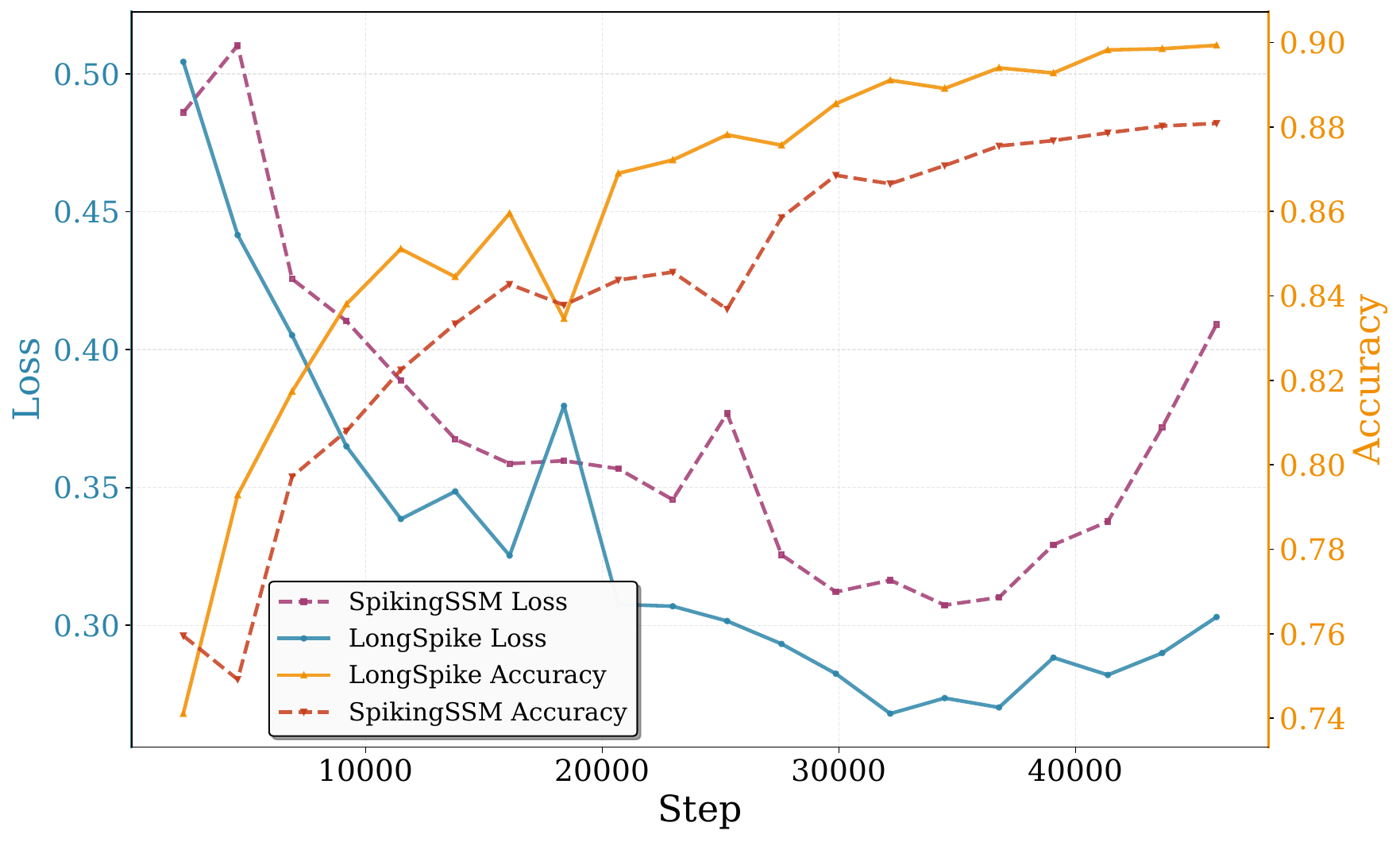}
\captionsetup{skip=2pt}
\captionof{figure}{Loss/accuracy trends on the LRA RETRIEVAL task.}
\label{metrics_plot}
\end{minipage}



In addition to the LRA benchmark, we evaluate LongSpike on the large-scale WikiText-103 dataset to assess its effectiveness on real-world language modeling tasks. 
As shown in Table~\ref{tab:wikitext_comparison}, LongSpike outperforms SpikingSSM with comparable model size, and achieves lower perplexity than SpikeGPT~\cite{zhu2023spikegpt} despite using significantly fewer parameters. 
Moreover, its spiking rate remains comparable to that of SpikingSSM, indicating that the performance gain is not obtained by substantially increased neural activity. 
These results further narrow the performance gap with ANN-based models, highlighting the advantage of fractional dynamics for long-context language modeling.


\begin{wraptable}[7]{r}{0.5\textwidth}   
\vspace{-20pt}                           
\centering
\setlength{\tabcolsep}{4pt}
\caption{Performance comparison on Speech Commands dataset.}
\label{tab:sc_comparison}
\begin{tabular}{lcc}
\toprule
Model & \#Params & Accuracy(\%) \\
\midrule
S4D-Lin & 306K & 96.25 \\
SpikingSSM & 305K  & 96.09 \\
\textbf{LongSpike} & 305K & \textbf{96.31} \\
\bottomrule
\end{tabular}
\end{wraptable}

Finally, we evaluate LongSpike on the Speech Commands dataset. As shown in Table~\ref{tab:sc_comparison}, it achieves a competitive accuracy of 96.31\%, surpassing the spiking baseline SpikingSSM. These results indicate that LongSpike can effectively model temporal dependencies in sequential audio data.

\subsection{Efficiency and Fractional Approximation Analysis}
\label{sec:efficiency_analysis}

A key feature of LongSpike is the use of a sum-of-exponentials approximation to implement fractional dynamics. When reduced to a single exponential ($M=1, \omega=0, \eta=1$) in Equation~\cref{eq:SOE}, corresponding to the kernel $k_\alpha(t)$ with $\alpha=1$, the model recovers the standard SpikingSSM~\cite{shen2025spikingssms}, yielding identical performance. Thus, LongSpike can be viewed as a fractional-order generalization of SpikingSSM, where increasing $M$ improves the approximation of the long-memory kernel at the cost of additional computation.

In our experiments, we use a small number of exponential terms ($M=2$) to balance modeling capacity and efficiency. This setting improves accuracy while preserving sparse neural activity. For example, on the TEXT task of the LRA benchmark, LongSpike achieves 88.19\% accuracy, substantially outperforming SpikingSSM with 80.41\%. Meanwhile, the spiking rate remains low (6.46\% vs.\ 6.75\%), indicating that the performance gain does not come from substantially denser spiking activity, but from the enhanced long-memory dynamics introduced by the fractional kernel.

From a computational perspective, LongSpike preserves the efficient SSM computation pattern. SpikingSSM has a forward-pass complexity of $O(HNL + BHL\log L)$, dominated by FFT-based convolution, while LongSpike only increases kernel construction from $O(HNL)$ to $O(MHNL)$ and leaves the FFT convolution unchanged. For small $M$ (e.g., $M=2$), this introduces only a constant-factor overhead; moreover, the memory complexity can be kept at $O(HNL)$ with a loop-based implementation, matching SpikingSSM. In practice, the computation can be efficiently parallelized on GPUs with CUDA-level optimizations, reducing intermediate memory materialization and runtime overhead. A full analysis of time complexity, space complexity, and spiking-rate-based energy proxy is provided in Appendix~\ref{sec:efficiency}.

\vspace{-0.15cm}
\section{Conclusion}
\vspace{-0.15cm}
In this work, we introduced LongSpike, a novel spiking neural network framework designed to overcome the ``memoryless'' bottleneck of traditional first-order SNNs. By integrating fractional-order state-space modeling ($f$-SSM) into the spiking domain, we have shifted the paradigm from Markovian state transitions to dynamics that inherently capture long-range temporal dependencies through power-law memory kernels. Our approach successfully addresses the historical computational challenges of fractional operators. By reformulating these dynamics into a parallelizable state-space representation, LongSpike achieves the best of both worlds: the biological plausibility and energy efficiency of SNNs, and the high-performance parallel training capabilities of modern SSMs. Empirical results across LRA, WikiText-103, and Speech Commands benchmarks demonstrate that LongSpike not only outperforms state-of-the-art SNNs in accuracy but also preserves sparse synaptic computation in modeling sequences with fine-grained, long-term temporal structures.

\vspace{-0.15cm}
\section{Limitations and Future Work}
\vspace{-0.15cm}
\label{sec:Limitation and Future Work}
We have not yet deployed LongSpike on dedicated neuromorphic hardware platforms such as Intel Loihi, as our current work mainly focuses on the algorithmic and modeling aspects of f-SSM. Nevertheless, LongSpike remains naturally compatible with neuromorphic deployment. During inference, the model operates entirely through recurrent state updates without dense sequence-level matrix multiplications, which is favorable for hardware-efficient computation. 
We provide a spiking-rate-based energy analysis in Appendix~\ref{sec:appendix_energy}; future work will investigate hardware-level implementations and optimized sparse approximations of the fractional kernel to further improve efficiency and evaluate real-world latency and energy consumption. 






\nocite{*}
\bibliographystyle{unsrtnat}
\bibliography{reference}


\appendix

\section*{Impact Statement}
\label{sec:Impact Statement}
This research contributes to the field of sequence modeling by introducing \emph{f}-SSM into SNNs. The positive impacts of this work include the development of more robust and energy-efficient models for speech recognition, time-series forecasting, and physical system modeling. By improving the memory capacity of SNNs without incurring quadratic computational costs, we enable the deployment of sophisticated temporal models in real-time safety-critical systems. We acknowledge that improvements in sequence modeling can be applied to dual-use technologies, such as deepfake generation or automated surveillance. However, our work primarily targets the efficiency bottleneck of current AI systems, aiming to reduce the computational resources required for state-of-the-art performance.

\section{More Technical Details}
\subsection{More About Fractional Calculus}

This section provides additional mathematical background for readers less familiar with fractional calculus; it is not required for understanding the main method.

In \cref{sec:Fractional Order Calculus} of the main paper, we introduced the left Caputo fractional derivative and discussed numerical schemes for solving fractional-order ordinary differential equations (ODEs). Here we provide additional background on fractional calculus underlying our approach.

It is worth emphasizing that fractional derivatives are not uniquely defined; multiple formulations exist with different analytical properties. In the main paper, unless otherwise specified, we adopt the left-sided Caputo fractional derivative, denoted by \( D^\alpha \), due to its suitability for modeling physical systems and compatibility with standard initial conditions.

\subsubsection{Classical Derivatives and Integrals}

For a scalar function \( y(t) \), the first-order derivative describes its instantaneous rate of change:
\begin{equation}
\frac{dy(t)}{dt} = \dot{y}(t) = \lim_{\Delta t \to 0} \frac{y(t + \Delta t) - y(t)}{\Delta t}.
\end{equation}

Let \( J \) denote the integral operator that maps a Riemann-integrable function \( y(t) \) on \( [0,T] \) to its antiderivative starting from \( a \):
\begin{equation}
J y(t) := \int_a^t y(\tau)\, d\tau, \quad t \in [0, T].
\end{equation}

For a positive integer \( n \in \mathbb{N}^+ \), we define the \( n \)-fold integral operator recursively as \( J^n := J \circ J^{n-1} \)for \( n \geq 2 \), and \( J^1 = J \). It can be shown via repeated integration by parts \cite{kai2004analysis} that:
\begin{equation}
\label{eq:integral1}
J^n y(t) = \frac{1}{(n-1)!} \int_a^t (t - \tau)^{n-1} y(\tau) d\tau.
\end{equation}

\subsubsection{Fractional Integrals}

Fractional calculus generalizes classical integration by extending the order of integration from integers to real (or even complex) values. Among various definitions, the Riemann--Liouville (RL) fractional integral is one of the most fundamental \cite{tarasov2011fractional}.

For \( \alpha \in \mathbb{R}^+ \), the left- and right-sided RL fractional integrals are defined as:
\begin{equation}
\label{eq:f_1}
J^\alpha_{\text{left}} y(t) = \frac{1}{\Gamma(\alpha)} \int_a^t (t - \tau)^{\alpha - 1} y(\tau)\, d\tau,
\end{equation}
\begin{equation}
\label{eq:f_2}
J^\alpha_{\text{right}} y(t) = \frac{1}{\Gamma(\alpha)} \int_t^b (\tau - t)^{\alpha - 1} y(\tau)\, d\tau,
\end{equation}
where \( \Gamma(\alpha) \) is the Gamma function, which generalizes the factorial to non-integer arguments.

Compared with \cref{eq:integral1}, fractional integrals allow the order parameter to vary continuously, providing a smooth interpolation between different levels of accumulation and enabling richer modeling flexibility.

\subsubsection{Fractional Derivatives}

Extending differentiation to non-integer orders is more subtle, and several definitions exist. Among them, the Riemann--Liouville (RL) and Caputo derivatives are the most commonly used.

\paragraph{Riemann--Liouville fractional derivative.}
The RL fractional derivative is defined by applying an integer-order derivative after a fractional integral:
\begin{equation}
\label{fd_1}
{}^{\mathrm{RL}}D^\alpha y(t)
:=
\frac{d^m}{dt^m}
J_{\mathrm{left}}^{m-\alpha} y(t)
=
\frac{1}{\Gamma(m-\alpha)}
\frac{d^m}{dt^m}
\int_a^t
\frac{y(\tau)\, d\tau}{(t-\tau)^{\alpha-m+1}},
\end{equation}

\begin{equation}
\label{fd_2}
{}^{\mathrm{RL}}D^\alpha_{b-} y(t)
:=
(-1)^m
\frac{d^m}{dt^m}
J_{\mathrm{right}}^{m-\alpha} y(t)
=
\frac{(-1)^m}{\Gamma(m-\alpha)}
\frac{d^m}{dt^m}
\int_t^b
\frac{y(\tau)\, d\tau}{(\tau-t)^{\alpha-m+1}}.
\end{equation}
where \( m \) is the smallest integer satisfies \( m-1 < \alpha \leq m \).

\paragraph{Caputo fractional derivative.}
The Caputo derivative reverses the order of operations: it first computes an integer-order derivative and then applies a fractional integral:
\begin{equation}
\label{fd_3}
D^\alpha y(t)
:=
J_{\mathrm{left}}^{m-\alpha}
\frac{d^m}{dt^m} y(t)
=
\frac{1}{\Gamma(m-\alpha)}
\int_a^t
\frac{\frac{d^m}{d\tau^m}y(\tau)\, d\tau}
{(t-\tau)^{\alpha-m+1}},
\end{equation}

\begin{equation}
\label{fd_4}
D^\alpha_{b-} y(t)
:=
(-1)^m
J_{\mathrm{right}}^{m-\alpha}
\frac{d^m}{dt^m} y(t)
=
\frac{(-1)^m}{\Gamma(m-\alpha)}
\int_t^b
\frac{\frac{d^m}{d\tau^m}y(\tau)\, d\tau}
{(\tau-t)^{\alpha-m+1}}.
\end{equation}

Compared with the RL formulation, the Caputo derivative has two important advantages:  
(i) it yields zero when applied to constant functions, consistent with classical derivatives;  
(ii) it allows the use of standard integer-order initial conditions in differential equations.  

For these reasons, the Caputo formulation is more commonly adopted in practical modeling and is used throughout this work.

\subsubsection{Memory Effect and Non-locality}

A key feature of fractional derivatives is their intrinsic non-locality. As shown in \cref{fd_1,fd_2,fd_3,fd_4}, the derivative at time \( t \) depends on the entire history of the function over an interval (e.g., \( [a, t] \) for left-sided operators). This stands in contrast to classical derivatives, which are purely local.

The kernel \( (t - \tau)^{-\alpha} \) determines how past states contribute to the present, with the fractional order \( \alpha \) controlling the decay rate of historical influence. This naturally introduces a form of long-range memory, which is particularly suitable for modeling systems with hereditary effects, long-term dependencies, or anomalous diffusion.

When the order \( \alpha \) becomes an integer \( n \in \mathbb{N} \), fractional operators recover their classical integer-order counterparts under appropriate regularity conditions \cite{kai2004analysis}. This ensures that fractional calculus forms a consistent generalization of standard calculus.

For vector-valued functions, fractional operators act component-wise, analogous to classical differentiation and integration.

\subsection{Traditional Numerical Methods for Solving FDEs}
\label{sec:numerical_methods}
Considering the problem given by \cref{eq:FDE} where $y_0$ is the given initial condition, we define $h > 0$ as the discretization parameter (step size) and create a uniform grid over the interval $[0, T]$ given by $\left\{t_j=a+j h: j=0,1, \ldots, N\right\}$, where $N = T/h$ represents the number of time steps.
\subsubsection{Fractional ABM predictor:}
The numerical method is based on the Volterra integral representation of \cref{eq:FDE}, where the $D^\alpha$ takes the Caputo Derivative:
\begin{equation}
    y(t)=y_0+\frac{1}{\Gamma(\alpha)}\int_0^t(t-\tau)^{\alpha-1}f(\tau,y(\tau))ds
\end{equation}
The fractional Adams-Bashforth predictor \cite{kai2004analysis} is given by the following iterations:
\begin{equation}
    y_{k+1}=y_0+\frac{1}{\Gamma(\alpha)}\sum _{j=0}^{k}b_{j,k+1}f(t_j,y_j)
\end{equation}
with weights 
\begin{equation}
    b_{j,k+1}=\frac{h^{\alpha}}{\alpha}((k+1-j)^\alpha-(k-j)^\alpha),\quad j=0,1, \ldots, k
\end{equation}
\begin{itemize}[label=$\bullet$, topsep=0pt, itemsep=3pt, partopsep=0pt, parsep=0pt,leftmargin=20pt]
\item \textbf{short memory version:} Given a memory length $M$ and we can truncate the convolution tail and get the short memory version:
\begin{equation}
    y_{k+1}=y_0+\frac{1}{\Gamma(\alpha)}\sum _{j=max\{0,k-M+1\}}^{k}b_{j,k+1}f(t_j,y_j)
\end{equation}
\end{itemize}

\subsection{\texorpdfstring{Methods for Generating $\{ \omega_i, \eta_i \}_{i=1}^M$}{Methods for Generating omega and eta parameters}}
\label{sec:Methods for Generating omega&eta}
Here we provide two methods, actually the $\{ \omega_i, \eta_i \}_{i=1}^M$ can be learnable.
\subsubsection{Log-Trapezoid Quadrature for the Diffusive Integral}
The first method uses log-trapezoidal quadrature to approximate the diffusive integral. Starting from the exact "diffusive" representation in \cref{eq:diffusive representation}, we let $\omega=e^x$, then the \cref{eq:diffusive representation} becomes:
\begin{equation}
    \frac{t^{\alpha - 1}}{\Gamma(\alpha)} = \frac{\sin(\pi \alpha)}{\pi} \int_{-\infty}^\infty e^{-te^x} e^{(1-\alpha)x} dx
\end{equation}
Then we truncate the frequency range $\omega \in [\omega_{\text{min}}, \omega_{\text{max}}]$ and apply the trapezoidal rule for numerical integration.
\begin{itemize}[label=$\bullet$, topsep=0pt, itemsep=3pt, partopsep=0pt, parsep=0pt,leftmargin=20pt]
\item\textbf{Choose frequency band:}  
Select a practical band that matches the time grid $[h, T]$:
\begin{equation}
    \omega_{\text{min}} = \frac{1}{T}, \quad \omega_{\text{max}} = \frac{c}{h}
\end{equation}
where $c$ is a safety factor (commonly chosen between 5 and 20, e.g., $c = 10$).\\

\item\textbf{Nodes and Weights (Closed Form):}  
The nodes $\omega_i$ and weights $\eta_i$ are determined as follows:
\begin{equation}
    x_i = x_{\text{min}} + (i - 1) \Delta x, \quad \omega_i = e^{x_i}, \quad \eta_i = \frac{\sin(\pi \alpha)}{\pi} \Delta x e^{(1-\alpha) x_i}=\frac{\sin(\pi \alpha)}{\pi} \Delta x \omega_i^{(1-\alpha)}
\end{equation}
with the trapezoidal half-weights:
\begin{equation}
    \eta_1 \leftarrow \frac{1}{2} \eta_1, \quad \eta_M \leftarrow \frac{1}{2} \eta_M
\end{equation}
\end{itemize}
\subsubsection{Log-Spaced Least Squares (Nonnegative)}
The second method uses log-spaced least squares to compute the weights $\eta_i$. This method is particularly effective for improving accuracy, especially for the same $M$.
\begin{itemize}[label=$\bullet$, topsep=0pt, itemsep=3pt, partopsep=0pt, parsep=0pt,leftmargin=20pt]
\item\textbf{Choose $\omega_i$:}  
We choose the same frequency band as the first method, but now the nodes $\omega_i$ are chosen on a logarithmic scale:
\begin{equation}
    \omega_i = \omega_{\text{min}} \left( \frac{\omega_{\text{max}}}{\omega_{\text{min}}} \right)^{\frac{i-1}{M-1}}, \quad i = 1, \dots, M
\end{equation}

\item\textbf{Fit $\eta$ on Sampled Times:}  
We pick $K$ sample times, and log-spaced in the interval $[h, T]$:
\begin{equation}
    t_n = h \left( \frac{T}{h} \right)^{\frac{n-1}{K-1}}, \quad n = 1, \dots, K
\end{equation}
Then build the matrix:
\begin{equation}
    A_{ni} = e^{-\omega_i t_n}, \quad b_n = \frac{t_n^{\alpha - 1}}{\Gamma(\alpha)}
\end{equation}
Finally solve for $\eta$ via non-negative least squares (NNLS):
\begin{equation}
    \min_{\eta \geq 0} \| A \eta - b \|_2^2
\end{equation}
Furthermore, we can also use the ridge regression:
\begin{equation}
    \eta = (A^T A + \lambda I)^{-1} A^T b, \quad \text{where} \quad \lambda > 0
\end{equation}
\end{itemize}

\section{Derivation of Standard and Fractional SSM Solutions}
\label{proof of Long-Memory Behavior}
\subsection{\texorpdfstring{Standard SSM ($\alpha=1$)}{Standard SSM (alpha = 1)}}
Consider the standard scalar SSM:
\begin{equation}
    \frac{dh(t)}{dt} = -\lambda h(t) + x_c, \quad h(0) = h_0
\end{equation}
with constant input $x_c$.

The homogeneous equation is
\begin{equation}
    \frac{dh_h(t)}{dt} = -\lambda h_h(t) \quad \Rightarrow \quad
h_h(t) = C e^{-\lambda t}.
\end{equation}
For constant input, assume $h_p = K$. Then
\begin{equation}
    0 = -\lambda K + x_c \quad \Rightarrow \quad K = \frac{x_c}{\lambda}.
\end{equation}

So the general solution is:
\begin{equation}
    h(t) = h_h(t) + h_p = C e^{-\lambda t} + \frac{x_c}{\lambda}.
\end{equation}
We apply initial condition $h(0) = h_0$ gives $C = h_0 - \frac{x_c}{\lambda}$. Therefore the solution is
\begin{equation}
    h(t) = \frac{x_c}{\lambda} + \left(h_0 - \frac{x_c}{\lambda}\right) e^{-\lambda t}.
\end{equation}
\subsection{Fractional SSM \texorpdfstring{$(0<\alpha<1)$}{(0 < alpha < 1)}}
Consider the fractional case:
\begin{equation}
    D^\alpha h(t) = -\lambda h(t) + x_c, \quad h(0) = h_0
\end{equation}
Using the Caputo derivative property:
\begin{equation}
    \mathcal{L}\{D^\alpha h(t)\} = s^\alpha H(s) - s^{\alpha-1} h_0
\end{equation}
we have
\begin{equation}
    s^\alpha H(s) - s^{\alpha-1} h_0 = -\lambda H(s) + \frac{x_c}{s}
\end{equation}
Then we solve for $H(s)$
\begin{equation}
    (s^\alpha + \lambda) H(s) = s^{\alpha-1} h_0 + \frac{x_c}{s} \quad \Rightarrow \quad
H(s) = \frac{s^{\alpha-1} h_0}{s^\alpha + \lambda} + \frac{x_c}{s(s^\alpha + \lambda)}.
\end{equation}
Using the Mittag-Leffler function property:
\begin{equation}
    \mathcal{L}^{-1}\left\{\frac{s^{\alpha-1}}{s^\alpha + \lambda}\right\} = E_\alpha(-\lambda t^\alpha),
\end{equation}
and
\begin{equation}
    \frac{1}{s(s^\alpha + \lambda)} = \frac{1}{\lambda} \left(\frac{1}{s} - \frac{s^{\alpha-1}}{s^\alpha + \lambda}\right),
\end{equation}
we obtain
\begin{equation}
    h(t) = h_0 E_\alpha(-\lambda t^\alpha) + \frac{x_c}{\lambda}\left[1 - E_\alpha(-\lambda t^\alpha)\right] = \frac{x_c}{\lambda} + \left(h_0 - \frac{x_c}{\lambda}\right) E_\alpha(-\lambda t^\alpha).
\end{equation}

\section{Experimental Details}
\label{sec:Experimental Settings}
The specific parameter settings for all experiments are summarized in \cref{tab:hyperparameters}. These settings cover six tasks from the Long Range Arena (LRA) benchmark, as well as experiments on sMNIST, psMNIST, WikiText-103, Speech Commands and DVS128-Gesture.

The LongSpike model utilizes the piecewise quadratic surrogate spiking function. We chose the piecewise quadratic surrogate because it balances stability, efficiency, and sparsity in SNN training: Its gradient is bounded and non-zero only in the local neighborhood near the threshold, which ensures stable backpropagation while avoiding noise updates from distant threshold states. It is computationally efficient and has a symmetric shape, which is also compatible with learnable thresholds. The gradient of this function is defined as:
\begin{equation}
    g'(x) = 
  \begin{cases} 
    0, & \text{if } |x| > \frac{1}{\alpha} \\
    -\alpha^2 |x| + \alpha, & \text{if } |x| \leq \frac{1}{\alpha}
  \end{cases}
\end{equation}
The primitive function is defined as:
\begin{equation}
    g(x) = 
  \begin{cases} 
    0, & \text{if } x < -\frac{1}{\alpha} \\
    -\frac{1}{2} \alpha^2 |x|x + \alpha x + \frac{1}{2}, & \text{if } |x| \leq \frac{1}{\alpha} \\
    1, & \text{if } x > \frac{1}{\alpha}
  \end{cases}
\end{equation}
In our experiments, the value of $\alpha$ is set to 1.

\begin{table*}[!htp]\small
\caption{The hyperparameters used in the experiments for various benchmark tasks. H denotes the model dimension, N denotes the state dimension, LR denotes learning rate, WD denotes weight decay, and BS denotes the batch size. BN and LN refer to Batch Normalization and Layer Normalization.}
\label{tab:hyperparameters}
\centering
\resizebox{\textwidth}{!}{
\setlength{\tabcolsep}{2pt} 
\begin{tabular}{lcccccccccccc}
\toprule
Dataset & Depth & H & N & Norm & pNorm & Dropout & LR & BS & Epochs & WD & $(\Delta_{\min}, \Delta_{\max})$ & Hardware \\
\midrule
ListOps    & 8 & 128 & 64 & BN & False & 0   & 0.01  & 50 & 40  & 0.05 & (0.001, 0.1) & 1×V100 \\
Text       & 6 & 256 & 64 & BN & True  & 0   & 0.01  & 16 & 32  & 0.01 & (0.001, 0.1) & 1×A100 \\
Retrieval  & 6 & 256 & 64 & BN & True  & 0   & 0.01  & 64 & 20  & 0.01 & (0.001, 0.1) & 1×A100 \\
Image      & 6 & 512 & 64 & LN & False & 0.1 & 0.01  & 50 & 200 & 0.01 & (0.001, 0.1) & 1×V100 \\
Pathfinder & 6 & 256 & 64 & BN & True  & 0   & 0.004 & 64 & 200 & 0.01 & (0.001, 0.1) & 1×V100 \\
Path-X     & 6 & 256 & 64 & BN & True  & 0   & 0.0005 & 32 & 50  & 0.01 & (0.0001, 0.01) & 1×A800 \\
\midrule
WT-103 &16& 1024& 64& LN& True &0.1& 0.0005& 1 & 200 & 0.01 &(0.001,0.1)& 8×H800\\
psMNIST    & 4 & 256 & 64 & LN & False & 0.1 & 0.01  & 50 & 100 & 0.01 & (0.001, 0.1) & 1×V100 \\
sMNIST     & 2 & 400 & 64 & LN & False & 0.1 & 0.01  & 50 & 100 & 0.01 & (0.001, 0.1) & 1×V100 \\
\midrule
DVS128 & 6 & 512 & 16 & LN & False & 0.1 & 0.005 & 32 & 200 & 0.05 & (0.001, 0.1) & 1×RTX 4090 \\
SC         & 6 & 128 & 32 & BN & True  & 0   & 0.01  & 16 & 40  & 0.01 & (0.001, 0.1) & 1×RTX 4090 \\
\bottomrule
\end{tabular}}
\vspace{-.3cm}
\end{table*}

Experiments are conducted on NVIDIA GPUs (e.g., NVIDIA V100, A100, A800, H800,and RTX 4090), selected according to the memory requirements of different tasks. Unless otherwise specified, all experiments are run on a single GPU.

\section{Computational Cost and Energy Analysis}
\label{sec:efficiency}

\subsection{Computational Complexity Analysis}
\label{sec:appendix_complexity}

We analyze the computational complexity of LongSpike relative to SpikingSSM, focusing on the common setting with two fractional terms ($M=2$).

\textbf{Time Complexity.}
SpikingSSM has a forward-pass complexity of $O(HNL + BHL \log L)$, which is dominated by FFT-based convolution.
LongSpike approximates the fractional kernel using $M$ exponential terms, increasing the kernel construction cost from $O(HNL)$ to $O(MHNL)$, while leaving the FFT-based convolution unchanged.
Furthermore, on GPU hardware, we can compute the effective kernel $K_{\text{eff}}$ in parallel via aggregation prior to convolution.
Overall, the additional cost introduces only a constant-factor overhead and preserves the same asymptotic dependence on sequence length.

\textbf{Space Complexity.}
LongSpike supports both vectorized and loop-based implementations.
The vectorized implementation requires $O(MHNL)$ memory, which remains acceptable for small $M$ (e.g., $M=2$), while the loop-based implementation reduces memory usage to $O(HNL)$, matching that of SpikingSSM.

In practice, we observe that LongSpike ($M=2$) introduces only a modest runtime overhead. 
As shown in Table~\ref{tab:sc_efficiency}, the per-epoch training time increases slightly, 
while GPU memory usage is reduced.

\begin{table}[H]
\centering
\caption{Training time and GPU memory usage on the Speech Commands dataset.}
\label{tab:sc_efficiency}
\begin{small}
\setlength{\tabcolsep}{5pt}
\begin{tabular}{lcc}
\toprule
Model & Training Time (Per-epoch)  & GPU Memory (MiB)  \\
\midrule
SpikingSSM & $\sim$26.5 min & 14429  \\
LongSpike ($M=2$) & $\sim$27.5 min & 12927  \\
Relative Change & $\uparrow$$\sim$3.8\% & $\downarrow$$\sim$10\%\\
\bottomrule
\end{tabular}
\end{small}
\end{table}

This reduction in memory footprint can be attributed to implementation-level optimizations, including CUDA-based kernel fusion and memory-efficient aggregation, which avoid materializing intermediate states. This suggests that fractional modeling does not inherently increase memory cost, and can even improve efficiency under optimized implementations.

\subsection{Energy Analysis via Spiking Activity}
\label{sec:appendix_energy}

Spiking neural networks (SNNs) are widely regarded as energy-efficient due to their sparse and binary activation patterns. Unlike artificial neural networks (ANNs), where the dominant operation is multiply-and-accumulate (MAC), SNNs primarily rely on synaptic accumulation (AC). In particular, binary spike activations allow multiplication with weights to be simplified to addition operations on certain neuromorphic hardware, leading to significantly lower energy consumption\cite{ICLR2024_e9882f7f}.

Although the hardware specifics and neuron dynamics are not considered here, a theoretical analysis can provide an estimate of SNN efficiency. Following prior work\cite{ICLR2024_e9882f7f,li2024efficient}, we assume the energy cost of a MAC operation to be $E_{\text{MAC}} = 4.6$ pJ and that of an AC operation to be $E_{\text{AC}} = 0.9$ pJ\cite{horowitz20141}.

We define the \emph{spiking rate} as the ratio of spike events to total neuron activations over time, and use the average spiking rate across all neurons as a proxy for network activity. Since the number of AC operations in SNNs is proportional to the spiking rate, this metric provides an estimate of effective computation and energy consumption.

\begin{figure}[H]
  \centering
  \includegraphics[width=0.95\columnwidth]{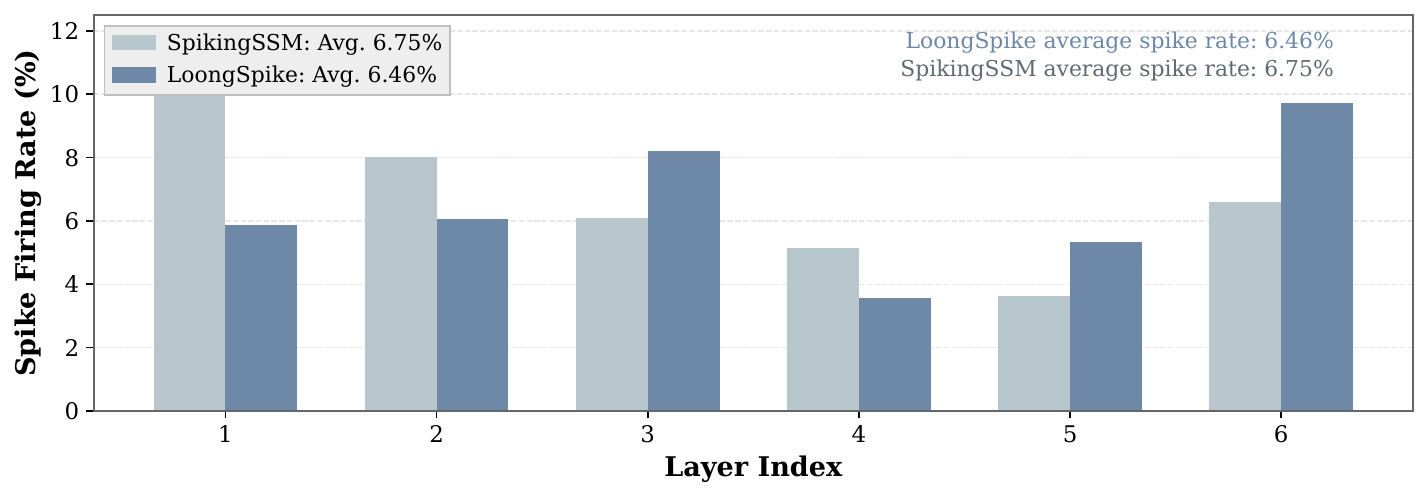}
  \caption{Layer-wise spiking rates on the TEXT task of the LRA benchmark. LongSpike maintains a comparable or lower average spiking rate (6.46\%) than SpikingSSM (6.75\%), indicating that performance improvements do not rely on increased neural activity.}
  \label{fig:spike_rate}
\end{figure}

In this analysis, we focus on the TEXT task of the LRA benchmark, where the model consists of 6 layers with a feature-mixing projection from $d=512$ to $d=1024$. Figure~\ref{fig:spike_rate} shows the layer-wise spiking rates. To quantify energy efficiency, we focus on the feature-mixing layers, which dominate computation, and estimate the number of operations and corresponding energy consumption based on these spiking statistics.

\begin{table}[h]
\centering
\caption{Energy-efficiency comparison on the TEXT task of the LRA benchmark.}
\label{tab:energy_efficiency}
\begin{small}
\setlength{\tabcolsep}{5pt}
\begin{tabular}{lcccc}
\toprule
Model & Op Type & Ops (G) & Energy (mJ) & Accuracy (\%) \\
\midrule
S4 (ANN) & MAC & 3.22 & 14.82 & 86.97 \\
SpikingSSM & AC & 0.217 & 0.196 & 80.41 \\
LongSpike & AC & 0.208 & 0.187 & \textbf{88.19} \\
\bottomrule
\end{tabular}
\end{small}
\end{table}

As shown in Table~\ref{tab:energy_efficiency}, both spiking models significantly reduce computation and energy consumption compared to the ANN baseline. Notably, LongSpike achieves higher accuracy while slightly reducing effective computation and energy compared to SpikingSSM. This indicates that the performance gains stem from improved long-range modeling rather than increased neural activity.

Overall, LongSpike preserves the energy-efficient characteristics of SNNs while enhancing sequence modeling capability. This analysis serves as a theoretical proxy for energy efficiency, since the actual energy consumption depends on hardware platforms and implementation details.

\section{Extended Experimental Results}
\label{sec:appendix_extra}
In this section, we present additional results that were not included in the main content.

\subsection{Sequential MNIST.}  
The MNIST dataset~\cite{lecun1998mnist} contains 70,000 grayscale images of handwritten digits (0–9), with 60,000 for training and 10,000 for testing, each of size 28×28 pixels. The Sequential MNIST (sMNIST)~\cite{le2015simple} flattens these images into sequences of 784 elements, while the Permuted Sequential MNIST (psMNIST)~\cite{le2015simple} applies a fixed pixel permutation, disrupting spatial correlations. These variants test a model's ability to learn sequence dependencies rather than relying on spatial patterns.

We evaluate LongSpike on the sMNIST and psMNIST datasets to examine its ability to model moderately long-range dependencies. As shown in Table~\ref{tab:mnist_comparison}, compared to state-of-the-art non-spiking (e.g., S4~\cite{gu2021efficiently}) and spiking models (e.g., SpikingSSM~\cite{shen2025spikingssms}), LongSpike achieves 99.62\% on sMNIST and 98.51\% on psMNIST, showing competitive performance on moderately long-range sequence tasks.

\begin{table}[!htp]
\caption{Performance comparison of LongSpike with other works on sMNIST and psMNIST datasets.}
\label{tab:mnist_comparison}
\centering
 \setlength{\tabcolsep}{2pt} 
 \begin{tabular}{lccc}
\toprule
Model & SNN & sMNIST& psMNIST \\
\midrule
LMUformer & No & 98.95 & 98.55 \\
 S4 & No & \textbf{99.63} & \textbf{98.70} \\
\midrule
SpikingLMUformer & Yes & 97.43 & 97.92 \\
Binary-S4D & Yes & 99.4 & - \\
P-SpikeSSM & Yes & 99.43 & 98.4 \\
SpikingSSM & Yes & 99.6 & 98.4 \\
\textbf{LongSpike} & Yes & \textbf{99.62} & \textbf{98.51} \\
\bottomrule
\end{tabular}
\vspace{-.3cm}
\end{table}

\subsection{DVS128 Gesture.}
The DVS128 Gesture dataset is a neuromorphic benchmark consisting of 11 hand gesture classes collected from 29 subjects under varying illumination conditions. The images are based on event-driven recordings, and we refer to previous work \cite{subramoney2022efficient} to preprocess the data format from event streams to frames. The results are reported in \cref{tab:dvs_comparison}.

\begin{table}[H]
\centering
\caption{Performance comparison of LongSpike and other works on the DVS128 Gesture dataset.}
\label{tab:dvs_comparison}
\begin{tabular}{lcc}
\toprule
    Model & \#Params & Accuracy(\%) \\
        \midrule
          SpikingSSM \cite{shen2025spikingssms} & 3.0M  &  97.1 \\
          LongSpike & 3.0M &  97.4 \\
    \bottomrule
\end{tabular}
\end{table}

\subsection{Effect of Fractional Dynamics without Spiking}

To further investigate the contribution of fractional dynamics independently of the spiking mechanism, we evaluate a non-spiking variant based on a vanilla fractional state space model (f-SSM).

Specifically, we replace the integer-order state space model (S4) with its fractional counterpart while removing all spiking components. This allows us to directly assess the impact of fractional-order dynamics on long-sequence modeling performance.

As shown in Table~\ref{tab:frac_ssm}, the vanilla f-SSM consistently outperforms the standard S4 model across all LRA tasks. These results suggest that the performance gains mainly stem from the improved long-memory modeling capability introduced by fractional dynamics, rather than the spiking mechanism itself.

\begin{table*}[!htp]\small
\caption{Comparison between integer-order SSM (S4) and fractional SSM (f-SSM) on LRA benchmark.}
\label{tab:frac_ssm}
\centering
 \resizebox{\textwidth}{!}{
 \setlength{\tabcolsep}{2pt} 
 \begin{tabular}{lcccccccccc}
\toprule
MODEL & SNN & LISTOPS & TEXT & RETRIEVAL & IMAGE & PATHFINDER & PATH-X & AVG \\
\midrule
S4 & No & 58.35 & 76.02 & 87.09 & 87.26 & 86.05 & 88.10 & 80.48 \\
f-SSM & No & 63.20 & 90.82 & 91.96 & 90.28 & 95.87 & 96.21 & 88.06 \\
\bottomrule
\end{tabular}}
\end{table*}


\newpage
\section*{NeurIPS Paper Checklist}

\begin{enumerate}

\item {\bf Claims}
    \item[] Question: Do the main claims made in the abstract and introduction accurately reflect the paper's contributions and scope?
    \item[] Answer: \answerYes{} 
    \item[] Justification: {The paper's contributions and scope are detailed in the abstract and introduction \cref{sec:introduction}.}
    \item[] Guidelines:
    \begin{itemize}
        \item The answer \answerNA{} means that the abstract and introduction do not include the claims made in the paper.
        \item The abstract and/or introduction should clearly state the claims made, including the contributions made in the paper and important assumptions and limitations. A \answerNo{} or \answerNA{} answer to this question will not be perceived well by the reviewers. 
        \item The claims made should match theoretical and experimental results, and reflect how much the results can be expected to generalize to other settings. 
        \item It is fine to include aspirational goals as motivation as long as it is clear that these goals are not attained by the paper. 
    \end{itemize}

\item {\bf Limitations}
    \item[] Question: Does the paper discuss the limitations of the work performed by the authors?
    \item[] Answer: \answerYes{} 
    \item[] Justification: {We have already discussed limitations in ~\cref{sec:Limitation and Future Work}.}
    \item[] Guidelines:
    \begin{itemize}
        \item The answer \answerNA{} means that the paper has no limitation while the answer \answerNo{} means that the paper has limitations, but those are not discussed in the paper. 
        \item The authors are encouraged to create a separate ``Limitations'' section in their paper.
        \item The paper should point out any strong assumptions and how robust the results are to violations of these assumptions (e.g., independence assumptions, noiseless settings, model well-specification, asymptotic approximations only holding locally). The authors should reflect on how these assumptions might be violated in practice and what the implications would be.
        \item The authors should reflect on the scope of the claims made, e.g., if the approach was only tested on a few datasets or with a few runs. In general, empirical results often depend on implicit assumptions, which should be articulated.
        \item The authors should reflect on the factors that influence the performance of the approach. For example, a facial recognition algorithm may perform poorly when image resolution is low or images are taken in low lighting. Or a speech-to-text system might not be used reliably to provide closed captions for online lectures because it fails to handle technical jargon.
        \item The authors should discuss the computational efficiency of the proposed algorithms and how they scale with dataset size.
        \item If applicable, the authors should discuss possible limitations of their approach to address problems of privacy and fairness.
        \item While the authors might fear that complete honesty about limitations might be used by reviewers as grounds for rejection, a worse outcome might be that reviewers discover limitations that aren't acknowledged in the paper. The authors should use their best judgment and recognize that individual actions in favor of transparency play an important role in developing norms that preserve the integrity of the community. Reviewers will be specifically instructed to not penalize honesty concerning limitations.
    \end{itemize}

\item {\bf Theory assumptions and proofs}
    \item[] Question: For each theoretical result, does the paper provide the full set of assumptions and a complete (and correct) proof?
    \item[] Answer: \answerYes{} 
    \item[] Justification: {The theorems proposed in this paper are supported by detailed proofs.}
    \item[] Guidelines:
    \begin{itemize}
        \item The answer \answerNA{} means that the paper does not include theoretical results. 
        \item All the theorems, formulas, and proofs in the paper should be numbered and cross-referenced.
        \item All assumptions should be clearly stated or referenced in the statement of any theorems.
        \item The proofs can either appear in the main paper or the supplemental material, but if they appear in the supplemental material, the authors are encouraged to provide a short proof sketch to provide intuition. 
        \item Inversely, any informal proof provided in the core of the paper should be complemented by formal proofs provided in appendix or supplemental material.
        \item Theorems and Lemmas that the proof relies upon should be properly referenced. 
    \end{itemize}

    \item {\bf Experimental result reproducibility}
    \item[] Question: Does the paper fully disclose all the information needed to reproduce the main experimental results of the paper to the extent that it affects the main claims and/or conclusions of the paper (regardless of whether the code and data are provided or not)?
    \item[] Answer: \answerYes{} 
    \item[] Justification: {We provide sufficient details on the model architecture, datasets, and training procedures to enable reproducibility of the reported results.}
    \item[] Guidelines:
    \begin{itemize}
        \item The answer \answerNA{} means that the paper does not include experiments.
        \item If the paper includes experiments, a \answerNo{} answer to this question will not be perceived well by the reviewers: Making the paper reproducible is important, regardless of whether the code and data are provided or not.
        \item If the contribution is a dataset and\slash or model, the authors should describe the steps taken to make their results reproducible or verifiable. 
        \item Depending on the contribution, reproducibility can be accomplished in various ways. For example, if the contribution is a novel architecture, describing the architecture fully might suffice, or if the contribution is a specific model and empirical evaluation, it may be necessary to either make it possible for others to replicate the model with the same dataset, or provide access to the model. In general. releasing code and data is often one good way to accomplish this, but reproducibility can also be provided via detailed instructions for how to replicate the results, access to a hosted model (e.g., in the case of a large language model), releasing of a model checkpoint, or other means that are appropriate to the research performed.
        \item While NeurIPS does not require releasing code, the conference does require all submissions to provide some reasonable avenue for reproducibility, which may depend on the nature of the contribution. For example
        \begin{enumerate}
            \item If the contribution is primarily a new algorithm, the paper should make it clear how to reproduce that algorithm.
            \item If the contribution is primarily a new model architecture, the paper should describe the architecture clearly and fully.
            \item If the contribution is a new model (e.g., a large language model), then there should either be a way to access this model for reproducing the results or a way to reproduce the model (e.g., with an open-source dataset or instructions for how to construct the dataset).
            \item We recognize that reproducibility may be tricky in some cases, in which case authors are welcome to describe the particular way they provide for reproducibility. In the case of closed-source models, it may be that access to the model is limited in some way (e.g., to registered users), but it should be possible for other researchers to have some path to reproducing or verifying the results.
        \end{enumerate}
    \end{itemize}

\item {\bf Open access to data and code}
    \item[] Question: Does the paper provide open access to the data and code, with sufficient instructions to faithfully reproduce the main experimental results, as described in supplemental material?
    \item[] Answer: \answerYes{} 
    \item[] Justification: {We have provided the source code in the supplementary material.}
    \item[] Guidelines:
    \begin{itemize}
        \item The answer \answerNA{} means that paper does not include experiments requiring code.
        \item Please see the NeurIPS code and data submission guidelines (\url{https://neurips.cc/public/guides/CodeSubmissionPolicy}) for more details.
        \item While we encourage the release of code and data, we understand that this might not be possible, so \answerNo{} is an acceptable answer. Papers cannot be rejected simply for not including code, unless this is central to the contribution (e.g., for a new open-source benchmark).
        \item The instructions should contain the exact command and environment needed to run to reproduce the results. See the NeurIPS code and data submission guidelines (\url{https://neurips.cc/public/guides/CodeSubmissionPolicy}) for more details.
        \item The authors should provide instructions on data access and preparation, including how to access the raw data, preprocessed data, intermediate data, and generated data, etc.
        \item The authors should provide scripts to reproduce all experimental results for the new proposed method and baselines. If only a subset of experiments are reproducible, they should state which ones are omitted from the script and why.
        \item At submission time, to preserve anonymity, the authors should release anonymized versions (if applicable).
        \item Providing as much information as possible in supplemental material (appended to the paper) is recommended, but including URLs to data and code is permitted.
    \end{itemize}

\item {\bf Experimental setting/details}
    \item[] Question: Does the paper specify all the training and test details (e.g., data splits, hyperparameters, how they were chosen, type of optimizer) necessary to understand the results?
    \item[] Answer: \answerYes{} 
    \item[] Justification: {We have included the implementation details in both the main text and the Appendix~\ref{sec:Experimental Settings}.}
    \item[] Guidelines:
    \begin{itemize}
        \item The answer \answerNA{} means that the paper does not include experiments.
        \item The experimental setting should be presented in the core of the paper to a level of detail that is necessary to appreciate the results and make sense of them.
        \item The full details can be provided either with the code, in appendix, or as supplemental material.
    \end{itemize}

\item {\bf Experiment statistical significance}
    \item[] Question: Does the paper report error bars suitably and correctly defined or other appropriate information about the statistical significance of the experiments?
    \item[] Answer: \answerNo{} 
    \item[] Justification: {We follow common practice in prior SSM and SNN literature by reporting results with a fixed random seed. Given the substantial computational cost of long-sequence benchmarks, experiments are conducted using standard single-run evaluation protocols. We observed consistent performance trends across preliminary repeated trials.}
    \item[] Guidelines:
    \begin{itemize}
        \item The answer \answerNA{} means that the paper does not include experiments.
        \item The authors should answer \answerYes{} if the results are accompanied by error bars, confidence intervals, or statistical significance tests, at least for the experiments that support the main claims of the paper.
        \item The factors of variability that the error bars are capturing should be clearly stated (for example, train/test split, initialization, random drawing of some parameter, or overall run with given experimental conditions).
        \item The method for calculating the error bars should be explained (closed form formula, call to a library function, bootstrap, etc.)
        \item The assumptions made should be given (e.g., Normally distributed errors).
        \item It should be clear whether the error bar is the standard deviation or the standard error of the mean.
        \item It is OK to report 1-sigma error bars, but one should state it. The authors should preferably report a 2-sigma error bar than state that they have a 96\% CI, if the hypothesis of Normality of errors is not verified.
        \item For asymmetric distributions, the authors should be careful not to show in tables or figures symmetric error bars that would yield results that are out of range (e.g., negative error rates).
        \item If error bars are reported in tables or plots, the authors should explain in the text how they were calculated and reference the corresponding figures or tables in the text.
    \end{itemize}

\item {\bf Experiments compute resources}
    \item[] Question: For each experiment, does the paper provide sufficient information on the computer resources (type of compute workers, memory, time of execution) needed to reproduce the experiments?
    \item[] Answer: \answerYes{} 
    \item[] Justification: {We report the computing resources in Appendix~\ref{sec:Experimental Settings}.}
    \item[] Guidelines:
    \begin{itemize}
        \item The answer \answerNA{} means that the paper does not include experiments.
        \item The paper should indicate the type of compute workers CPU or GPU, internal cluster, or cloud provider, including relevant memory and storage.
        \item The paper should provide the amount of compute required for each of the individual experimental runs as well as estimate the total compute. 
        \item The paper should disclose whether the full research project required more compute than the experiments reported in the paper (e.g., preliminary or failed experiments that didn't make it into the paper). 
    \end{itemize}
    
\item {\bf Code of ethics}
    \item[] Question: Does the research conducted in the paper conform, in every respect, with the NeurIPS Code of Ethics \url{https://neurips.cc/public/EthicsGuidelines}?
    \item[] Answer:\answerYes{} 
    \item[] Justification: {We conform, in every respect, with the NeurIPS Code of Ethics.}
    \item[] Guidelines:
    \begin{itemize}
        \item The answer \answerNA{} means that the authors have not reviewed the NeurIPS Code of Ethics.
        \item If the authors answer \answerNo, they should explain the special circumstances that require a deviation from the Code of Ethics.
        \item The authors should make sure to preserve anonymity (e.g., if there is a special consideration due to laws or regulations in their jurisdiction).
    \end{itemize}

\item {\bf Broader impacts}
    \item[] Question: Does the paper discuss both potential positive societal impacts and negative societal impacts of the work performed?
    \item[] Answer: \answerYes{} 
    \item[] Justification: {We have discussed broader impacts in Appendix~\ref{sec:Impact Statement}.}
    \item[] Guidelines:
    \begin{itemize}
        \item The answer \answerNA{} means that there is no societal impact of the work performed.
        \item If the authors answer \answerNA{} or \answerNo, they should explain why their work has no societal impact or why the paper does not address societal impact.
        \item Examples of negative societal impacts include potential malicious or unintended uses (e.g., disinformation, generating fake profiles, surveillance), fairness considerations (e.g., deployment of technologies that could make decisions that unfairly impact specific groups), privacy considerations, and security considerations.
        \item The conference expects that many papers will be foundational research and not tied to particular applications, let alone deployments. However, if there is a direct path to any negative applications, the authors should point it out. For example, it is legitimate to point out that an improvement in the quality of generative models could be used to generate Deepfakes for disinformation. On the other hand, it is not needed to point out that a generic algorithm for optimizing neural networks could enable people to train models that generate Deepfakes faster.
        \item The authors should consider possible harms that could arise when the technology is being used as intended and functioning correctly, harms that could arise when the technology is being used as intended but gives incorrect results, and harms following from (intentional or unintentional) misuse of the technology.
        \item If there are negative societal impacts, the authors could also discuss possible mitigation strategies (e.g., gated release of models, providing defenses in addition to attacks, mechanisms for monitoring misuse, mechanisms to monitor how a system learns from feedback over time, improving the efficiency and accessibility of ML).
    \end{itemize}
    
\item {\bf Safeguards}
    \item[] Question: Does the paper describe safeguards that have been put in place for responsible release of data or models that have a high risk for misuse (e.g., pre-trained language models, image generators, or scraped datasets)?
    \item[] Answer: \answerNA{} 
    \item[] Justification: {The paper poses no such risks.}
    \begin{itemize}
        \item The answer \answerNA{} means that the paper poses no such risks.
        \item Released models that have a high risk for misuse or dual-use should be released with necessary safeguards to allow for controlled use of the model, for example by requiring that users adhere to usage guidelines or restrictions to access the model or implementing safety filters. 
        \item Datasets that have been scraped from the Internet could pose safety risks. The authors should describe how they avoided releasing unsafe images.
        \item We recognize that providing effective safeguards is challenging, and many papers do not require this, but we encourage authors to take this into account and make a best faith effort.
    \end{itemize}

\item {\bf Licenses for existing assets}
    \item[] Question: Are the creators or original owners of assets (e.g., code, data, models), used in the paper, properly credited and are the license and terms of use explicitly mentioned and properly respected?
    \item[] Answer: \answerYes{} 
    \item[] Justification: {We cite the original paper that produced the code package or dataset.}
    \item[] Guidelines:
    \begin{itemize}
        \item The answer \answerNA{} means that the paper does not use existing assets.
        \item The authors should cite the original paper that produced the code package or dataset.
        \item The authors should state which version of the asset is used and, if possible, include a URL.
        \item The name of the license (e.g., CC-BY 4.0) should be included for each asset.
        \item For scraped data from a particular source (e.g., website), the copyright and terms of service of that source should be provided.
        \item If assets are released, the license, copyright information, and terms of use in the package should be provided. For popular datasets, \url{paperswithcode.com/datasets} has curated licenses for some datasets. Their licensing guide can help determine the license of a dataset.
        \item For existing datasets that are re-packaged, both the original license and the license of the derived asset (if it has changed) should be provided.
        \item If this information is not available online, the authors are encouraged to reach out to the asset's creators.
    \end{itemize}

\item {\bf New assets}
    \item[] Question: Are new assets introduced in the paper well documented and is the documentation provided alongside the assets?
    \item[] Answer: \answerNA{} 
    \item[] Justification: {The paper does not release new assets.}
    \item[] Guidelines:
    \begin{itemize}
        \item The answer \answerNA{} means that the paper does not release new assets.
        \item Researchers should communicate the details of the dataset\slash code\slash model as part of their submissions via structured templates. This includes details about training, license, limitations, etc. 
        \item The paper should discuss whether and how consent was obtained from people whose asset is used.
        \item At submission time, remember to anonymize your assets (if applicable). You can either create an anonymized URL or include an anonymized zip file.
    \end{itemize}

\item {\bf Crowdsourcing and research with human subjects}
    \item[] Question: For crowdsourcing experiments and research with human subjects, does the paper include the full text of instructions given to participants and screenshots, if applicable, as well as details about compensation (if any)? 
    \item[] Answer: \answerNA{} 
    \item[] Justification: {The paper does not involve crowdsourcing nor research with human subjects.}
    \item[] Guidelines:
    \begin{itemize}
        \item The answer \answerNA{} means that the paper does not involve crowdsourcing nor research with human subjects.
        \item Including this information in the supplemental material is fine, but if the main contribution of the paper involves human subjects, then as much detail as possible should be included in the main paper. 
        \item According to the NeurIPS Code of Ethics, workers involved in data collection, curation, or other labor should be paid at least the minimum wage in the country of the data collector. 
    \end{itemize}

\item {\bf Institutional review board (IRB) approvals or equivalent for research with human subjects}
    \item[] Question: Does the paper describe potential risks incurred by study participants, whether such risks were disclosed to the subjects, and whether Institutional Review Board (IRB) approvals (or an equivalent approval/review based on the requirements of your country or institution) were obtained?
    \item[] Answer: \answerNA{} 
    \item[] Justification: {The paper does not involve crowdsourcing nor research with human subjects.}
    \item[] Guidelines:
    \begin{itemize}
        \item The answer \answerNA{} means that the paper does not involve crowdsourcing nor research with human subjects.
        \item Depending on the country in which research is conducted, IRB approval (or equivalent) may be required for any human subjects research. If you obtained IRB approval, you should clearly state this in the paper. 
        \item We recognize that the procedures for this may vary significantly between institutions and locations, and we expect authors to adhere to the NeurIPS Code of Ethics and the guidelines for their institution. 
        \item For initial submissions, do not include any information that would break anonymity (if applicable), such as the institution conducting the review.
    \end{itemize}

\item {\bf Declaration of LLM usage}
    \item[] Question: Does the paper describe the usage of LLMs if it is an important, original, or non-standard component of the core methods in this research? Note that if the LLM is used only for writing, editing, or formatting purposes and does \emph{not} impact the core methodology, scientific rigor, or originality of the research, declaration is not required.
    \item[] Answer: \answerNA{}.
    \item[] Justification: LLMs are not used as an important, original, or non-standard component of the core method or experiments.
    \item[] Guidelines:
    \begin{itemize}
        \item The answer \answerNA{} means that the core method development in this research does not involve LLMs as any important, original, or non-standard components.
        \item Please refer to our LLM policy in the NeurIPS handbook for what should or should not be described.
    \end{itemize}

\end{enumerate}

\end{document}